\documentclass{article}

\usepackage[preprint]{neurips_2024}

\usepackage[utf8]{inputenc} 
\usepackage[T1]{fontenc}    
\usepackage[hidelinks]{hyperref}       
\usepackage{url}            
\usepackage{booktabs}       
\usepackage{amsfonts}       
\usepackage{nicefrac}       
\usepackage{microtype}      
\usepackage{xcolor}         
\usepackage{xspace}
\usepackage{natbib}

\usepackage{graphicx} 
\usepackage{amsmath}
\usepackage{multirow}
\usepackage{comment}

\definecolor{ggreen}{HTML}{00A64F}
\definecolor{light-gray}{gray}{0.9}

\newcommand*{\tightcolorbox}[2]{%
    \begingroup\setlength{\fboxsep}{1pt}%
        \colorbox{#1}{{\hspace*{2pt}\vphantom{Ay}#2\hspace*{2pt}}}%
    \endgroup
}
\newcommand*{\code}[1]{\tightcolorbox{light-gray}{\texttt{#1}}}
\newcommand*{\newterm}[1]{{\textbf{\textsc{#1}}}}
\newcommand*{\modelname}[1]{{\textsc{#1}}}
\newcommand*{\datasetname}[1]{{\textsc{#1}}}
\newcommand*{\ourmodel}{\modelname{SUMA}\xspace}

\usepackage{pifont}
\newcommand{\cmark}{\ding{51}}%
\newcommand{\xmark}{\ding{55}}%

\newcommand\blfootnote[1]{%
  \begingroup
  \renewcommand\thefootnote{}\footnote{#1}%
  \addtocounter{footnote}{-1}%
  \endgroup
}

\title{Brain-Like Language Processing via a \\ Shallow Untrained Multihead Attention Network}

\author{%
    Badr AlKhamissi$^1$
    \qquad 
    Greta Tuckute$^2$
    \qquad
    Antoine Bosselut\hspace{1pt}$^{*,1}$
    \qquad 
    Martin Schrimpf\hspace{1pt}$^{*,1}$
    \\
    \\
    $^1$EPFL \quad $^2$MIT
    \\
}

\begin{document}

\blfootnote{$^*$ Equal Supervision}

\maketitle

\begin{abstract}
    Large Language Models (LLMs) have been shown to be effective models of the human language system, with some models predicting most explainable variance of brain activity in current datasets. Even in untrained models, the representations induced by architectural priors can exhibit reasonable alignment to brain data. In this work, we investigate the key architectural components driving the surprising alignment of untrained models. To estimate LLM-to-brain similarity, we first select language-selective units within an LLM, similar to how neuroscientists identify the language network in the human brain. We then benchmark the brain alignment of these LLM units across five different brain recording datasets. By isolating critical components of the Transformer architecture, we identify tokenization strategy and multihead attention as the two major components driving brain alignment. A simple form of recurrence further improves alignment. We further demonstrate this quantitative brain alignment of our model by reproducing landmark studies in the language neuroscience field, showing that localized model units -- just like language voxels measured empirically in the human brain -- discriminate more reliably between lexical than syntactic differences, and exhibit similar response profiles under the same experimental conditions. Finally, we demonstrate the utility of our model's representations for language modeling, achieving improved sample and parameter efficiency over comparable architectures. Our model's estimates of surprisal sets a new state-of-the-art in the behavioral alignment to human reading times. Taken together, we propose a highly brain- and behaviorally-aligned model that conceptualizes the human language system as an untrained shallow feature encoder, with structural priors, combined with a trained decoder to achieve efficient and performant language processing.\footnote[1]{Code and data available at \url{https://github.com/bkhmsi/brain-language-suma}}

\end{abstract}

\section{Introduction}

\begin{figure}
    \centering
    \includegraphics[width=1\linewidth]{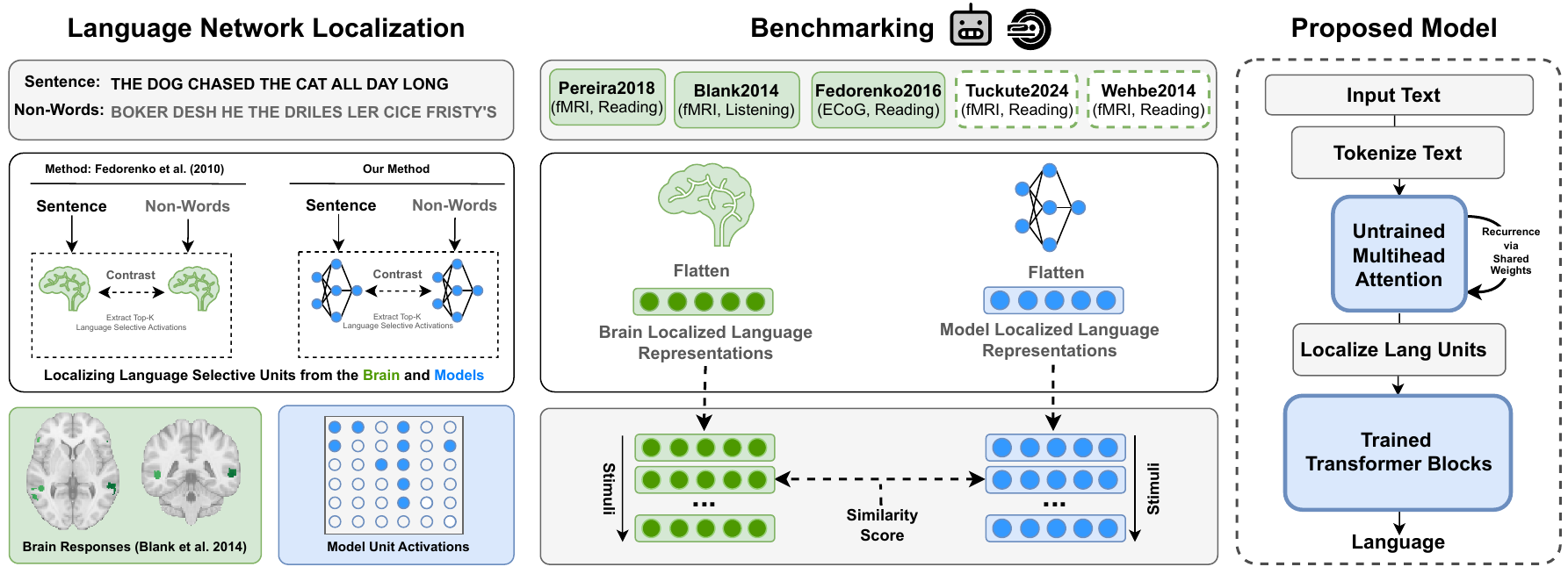}
    \caption{\textbf{Evaluating Model Alignment to the Human Language System.}
    \textbf{(Left) Localization: } We select the top-k language selective units in models and brain recordings by contrasting the difference in unit activations between sentences and lists of non-words, following \citet{Fedorenko2010NewMF}. \textbf{(Center) Benchmarking: } Across five different neural datasets, we measure the alignment between language selective units in models and the human brain. Each model's score is the mean of linear predictivity scores for each of the five datasets. We built our model using the first three benchmarks for validation and additionally report scores on two held-out benchmarks. \textbf{(Right) Proposed Model: } We conceptualize language processing in the human brain as an untrained feature encoder (\ourmodel) providing representations to a downstream trainable decoder that produces language output.}
    \label{fig:methods}
\end{figure}

Deciphering the brain's algorithms underlying our ability to process language and communicate is a core goal in neuroscience. Our ability to process language is supported by the human language system (HLS), a set of left-lateralized frontotemporal regions in the brain \citep[Figure \ref{fig:methods}]{Binder1997,Bates2003,GornoTempini2004,Price2010,Fedorenko2014,Hagoort2019} that respond robustly and selectively to linguistic input \citep{Fedorenko2024}. Driven by recent advances in machine learning, large language models (LLMs) trained via next-word prediction on large corpora of text, are now a particularly promising model family to capture the internal processes of the HLS. When exposed to the same text stimuli (e.g., sentences or narratives) as human participants during neuroimaging and electrophysiology sessions, certain LLMs predict most of the variance of human neural responses relative to the estimated reliability of the datasets \citep{schrimpf-pnas, caucheteux2022brains, pasquiou2022neural}. One surprising observation is that untrained models exhibit internal representations that are about half as brain-like as those of their trained counterparts \citep{schrimpf-pnas, pasquiou2022neural}. In this work, we investigate the reasons underlying the high brain alignment of untrained models, and, more broadly, isolate the critical model components that enable LLMs to capture human responses to language. 


To isolate the components driving model-to-brain alignment, we incrementally construct a single Transformer block from the ground up, beginning with static embeddings for each word and progressing to a complete single-block architecture. At each stage of this process, we measure brain alignment across three datasets \citep{pereira_toward_2018,fedorenko2016,blank_functional_2014} to guide our design choices and subsequently evaluate our final model on two held-out datasets \citep{tuckute2024driving,Wehbe2014}. 
Notably, we find that even without any training, the representations encoded by architectural priors alone are highly aligned to brain data.
We end-up with a simple untrained model that is able to explain most variance in current brain recording benchmarks. Beyond quantitative predictions, responses from this shallow untrained multihead attention architecture (\newterm{\ourmodel}) exhibit similar response profiles to those shown empirically in studies on the HLS \citep{Fedorenko2010NewMF, Fedorenko2012, Shain2023}, such as being more sensitive to lexical over syntactic differences. 
To test language production capabilities, we investigate whether this untrained model can serve as a good base for language modeling by feeding \ourmodel outputs to a trainable downstream decoder model. This combined model outperforms comparable trained architectures in terms of sample-efficiency and parameter-efficient perplexity and achieves state-of-the-art behavioral alignment on a benchmark \citep{futrell_natural_2018} predicting human reading times.
Taken together, our simple untrained model yields representations that are consistent with the human language system and capable of supporting downstream language production. Mapped to language processing in the brain, the human language system might thus serve as a simple generic feature encoder that provides representations for a downstream decoder. 

\section{Preliminaries \& Related Work}
\label{sec:preliminaries}

\paragraph{A Primer on Language in the Human Brain}
The Human Language System (HLS) is a set of brain regions supporting language that are functionally defined by their increased activity to language inputs over perceptually matched controls (e.g. lists of non-words) \citep[][Section \ref{sec:localization}]{Fedorenko2010NewMF}. These regions are typically lateralized to left-hemisphere frontal and temporal areas and exhibit remarkable selectivity for language processing compared to various non-linguistic inputs and tasks, such as music perception \citep{Fedorenko2012music, chen2023human} or arithmetic computation \citep{fedorenko2011functional} (for review, see \citep{Fedorenko2024}). These language regions show only weak responses when participants comprehend or articulate meaningless non-words \citep{Fedorenko2010NewMF,hu2023precision}. This selectivity profile is supported by extensive neuroimaging research and further corroborated by behavioral evidence from aphasia studies: when brain damage is confined to language areas, individuals lose their linguistic abilities while retaining other skills, such as mathematics \citep{Benn2013, Varley2005}, general reasoning \citep{Varley2000}, and theory of mind \citep{Siegal2006}. 



\paragraph{Model-to-Brain Alignment}
Previous studies have shown that the internal representations of certain artificial neural networks closely resemble those in the brain. This alignment was initially observed in the domain of vision \citep{yamins2014performance, cichy2016comparison, schrimpf_brain-score_2018, schrimpf2020, cadena2019deep, Kubilius2019} and has more recently been extended to language processing \citep{schrimpf-pnas, caucheteux2022brains, goldstein_shared_2022, Kauf23lexical, hosseini2024artificial, aw2023instructiontuning, tuckute2024driving}, and auditory processing \citep{kell2018task,Tuckute23many,koumura2023human}. Importantly, most of these models have been trained on large quantities of data. Within the domain of language, we here show that this is not necessarily required (see also \cite{hosseini2024artificial}).

\paragraph{Untrained Models}
Recent work has shown that an untrained convolutional network can yield high brain alignment to recordings in the visual ventral stream without the need for experience-dependent training \citep{Geiger2022,Kazemian2024}. Other works have investigated the inductive biases in different architectures and initializations in models of visual processing \citep{cichy2016comparison, cadena2019deep, Geiger2022}, speech perception \citep{Millet2021InductiveBP, Tuckute23many}, and language \citep{schrimpf-pnas, pasquiou2022neural}, highlighting that randomly initialized networks are not random functions \citep{teney2024neural}. We are not aware of any studies that thoroughly investigates the alignment of untrained models in language.

\paragraph{Spurious Correlations}
There have been concerns in recent studies about what aspects are captured by both trained and untrained language models when measuring brain alignment \citep{Kauf23lexical, antonello2024scaling, feghhi2024large}.
Therefore, to ensure our results are meaningful, we validate our findings by measuring alignment under several control conditions. Specifically, we observe a decrease in alignment on some datasets as we alter the input from the original sentence used in the neuroimaging study to the same sequence of tokens but shuffled and finally to a random sentence with the same sequence length. We performed these analyses across different metrics and datasets, and selected linear predictivity as our primary metric since it was the only one that best behaved as expected under the control conditions. We additionally replicated the control experiments done by \citet{feghhi2024large} on the \datasetname{Pereira2018} dataset and observe that simple properties such as sentence length and sentence position only predict a small portion of the variance relative to the models we tested (Appendix \ref{app:control-conditions-metrics}). 


\begin{table}[t]
    \centering
    \caption{
    \textbf{Datasets Used for Evaluating Model Alignment.} Neuroimaging datasets were collected via either functional magnetic resonance imaging (fMRI) or electrocorticography (ECoG). Text stimuli range from short sentences (\datasetname{Fedorenko2016}, \datasetname{Tuckute2024}) to paragraphs (\datasetname{Pereira2018}) and entire stories (\datasetname{Blank2014}, 
    \datasetname{Wehbe2014},
    \datasetname{Futrell2018}). \datasetname{Futrell2018} is a behavioral dataset.}
    \begin{tabular}{llll}
    \toprule 
    \textbf{Dataset} & \textbf{Modality} & \textbf{Stimulus Example} \\
    \midrule
    \textsc{Pereira2018} & fMRI (Reading) & Accordions produce sound with bellows ... \\
    \textsc{Blank2014} & fMRI (Listening) & A clear and joyous day it was and out on the wide ...   \\
    \textsc{Fedorenko2016} & ECoG (Reading) & `ALEX', `WAS', `TIRED', `SO', `HE', `TOOK', ... \\
    \textsc{Tuckute2024} & fMRI (Reading) & The judge spoke, breaking the silence. \\
    \textsc{Wehbe2014} & fMRI (Reading) & Harry had never believed he would meet a boy ... \\
    \midrule
    \textsc{Futrell2018} & Reading Times & A clear and joyous day it was and out on the wide ... \\
    \bottomrule \\
    \end{tabular}
    \label{tab:dataset}
\end{table}

\section{Localization of the Language Network}
\label{sec:localization}


The Human Language System (HLS) is defined\emph{ functionally} which means that units are chosen according to a `localizer' experiment \citep{saxe2006divide}. Specifically, the HLS is the set of neural units (e.g., voxels/electrodes) that are more selective to sentences over a perceptually-matched control condition \citep{Fedorenko2010NewMF}.
When selecting units from artificial models for comparison against HLS units, previous work selected output units from an entire Transformer block based on brain alignment scores \citep{schrimpf-pnas}. However, LLMs learn diverse concepts and behaviors during their considerable pretraining, not all of which are necessarily related to language processing, e.g., storage of knowledge \citep{AlKhamissi2022ARO} in the MLP layers \citep{geva-etal-2021-transformer} and ability to perform complex reasoning \citep{huang-chang-2023-towards}.
Therefore, we here characterize the language units in artificial neural networks using functional localization as is already standard in neuroscience. This approach comes with the advantage of comparability across different models since we can choose a fixed set of units which are localized independently of the critical experiment or modality. 



\begin{figure}[t]
    \centering
    \includegraphics[width=1\linewidth]{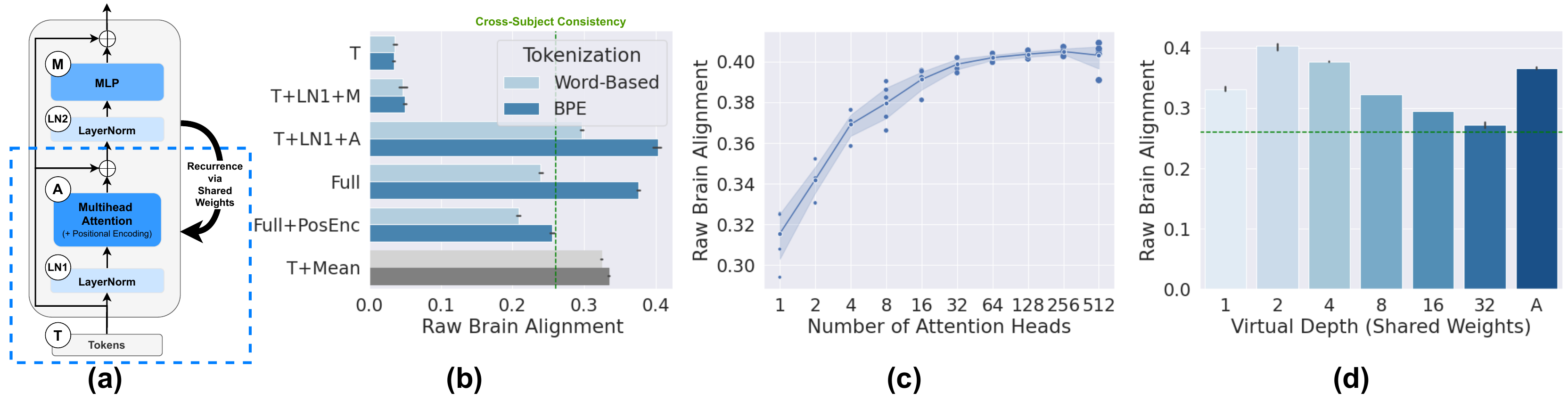}
    \caption{
        \textbf{Isolating Critical Components of the Transformer Architecture.} All models are untrained, i.e., representations are driven by architectural priors alone. Brain alignment is evaluated via ridge regression using each model's top 4096 language-selective units (Figure \ref{fig:methods}) on the three validation datasets. The {\color{ggreen} green} dashed line indicates the estimated data reliability (cross-subject consistency). Each experiment is repeated 5 times with different model seed initializations indicated by the error bars.
        \textbf{(a)} Transformer block with components labeled for the ablation study in (b). The {\color{blue} blue} dashed box indicates our final model \ourmodel.
        \textbf{(b)} Multihead Attention and tokenization strategy drives brain alignment for \ourmodel. Brain alignment of a single block model with different architectural ablations (labels as in (a)). Attention (\code{A}) is implemented with 512 attention heads. The virtual depth of the model is two layers. Representations are taken in response to the last token. \code{Mean} implies taking the average of all tokens. \code{PosEnc} implies using positional encoding.
        \textbf{(c)} Increasing the number of attention heads increases brain alignment. Base architecture is two layers of \code{T+LN1+A}.
        \textbf{(d)} Recurrent application of weights increases brain alignment. Base architecture is \code{T+LN1+A}. Virtual depth is increased by unrolling the same set of weights multiple times in the depth dimension (a simple form of recurrence). ``A'' refers to adaptive depth relative to the number of tokens measured using $\operatorname{ceiling}(\frac{\textnormal{\# of tokens}}{256})$.
    }
    \label{fig:ablations}
\end{figure}

Specifically, we present a set of sentences and strings of non-words of the same sequence length to each model, obtaining activations for each stimulus from units in the model at the output of different architectural components. We then define the model language system as the top-k units (with $k = 4096$ for all models to keep feature sizes comparable) that maximize the difference between activations to sentences and strings of non-words, measured using positive t-values with a Welch's t-test (Figure \ref{fig:methods}; see Appendix \ref{app:num-units} for the effect of different values for $k$ and Appendix \ref{app:localization-viz} for a visualization of the selected units for the pretrained and untrained versions of \modelname{GPT2-XL} and \modelname{LLaMA-2-7b}). This localization method selects a distributed set of units across the entire network, rather than restricting the representations to a single layer as done in prior work. We examine unit activations after 4 components in each Transformer block: (1) input layer normalization, (2) multihead self-attention, (3) post-attention layer normalization, and (4) the MLP (see Figure \ref{fig:ablations}(a) for Transformer architecture). For instance, for a model like \modelname{LLaMA-2-7B} \citep{llama2} which consists of 32 Transformer blocks and a hidden dimension of 4096, we consider $32 \times 4 \times 4096 = 524,288$ units, from which we select 4096 as the model's language system.

\section{Benchmarks}
\label{sec:benchmarks}

\paragraph{Datasets}

The neuroimaging datasets used in this work can be categorized along three dimensions: the imaging modality, the context length, and the modality through which the language stimulus was presented to human participants (auditory or visual). The imaging modalities we consider are functional magnetic resonance imaging (fMRI), which measures brain activity by detecting changes associated with blood oxygenation, and electrocorticography (ECoG) which records electrical activity via electrodes in direct contact with the brain. Table \ref{tab:dataset} provides an overview of all datasets in this study.
To investigate the robustness of our results, we evaluate the same neural datasets used by \citet{schrimpf-pnas}, and also test on the \datasetname{Wehbe2014} dataset \citep{Wehbe2014} and the distributionally-diverse \datasetname{Tuckute2024} dataset \citep{tuckute2024driving}.\footnote{Both datasets were only used until all design choices were fixed.} The datasets used to guide our design choices are from \cite{pereira_toward_2018, blank_functional_2014, fedorenko2016}, each covering different imaging modalities, context lengths, and stimulus presentation modality. We only consider neural units (electrodes, voxels, or regions) associated with the brain's language system that were localized by their respective authors using the method described in Section \ref{sec:localization}. However, in \datasetname{Wehbe2014}, functional localization was not performed -- to approximate the language regions, we extract the top-10\% voxels from each anatomically defined language region according to a probabilistic atlas for the human language system \citep{Lipkin2022}. In an additional analysis, we investigate alignment with language behavior using the \cite{futrell_natural_2018} dataset, which consists of self-paced per-word human reading times. See Appendix \ref{app:datasets} for details of each dataset.

\paragraph{Metrics}
Following standard practice in measuring brain alignment, we train a ridge regression model to predict brain activity from model representations, using the same input stimuli presented to human participants in neuroimaging studies \citep{Toneva2019InterpretingAI, tuckute2024driving}. We then measure the Pearson correlation between the predicted brain activations and the actual brain activations of human participants on a held-out set. This process is repeated over 10 cross-validation splits, and we report the average (mean) Pearson correlation as our final result. We refer to this metric as Linear Predictivity. Appendix \ref{app:add-metrics} shows results for non-parametric metrics: Centered Kernel Alignment (CKA; \citealp{cka}) and Representational Dissimilarity Matrices (RDM; \citealp{rdm}). 

\paragraph{Estimation of Cross-Subject Consistency}
To estimate the reliability of our datasets and the noise inherent in brain recordings, we compute each benchmark's \emph{cross-subject consistency}---referred to as the noise ceiling in previous work \citep{schrimpf-pnas}. 
For benchmarks with a Linear Predictivity metric, we estimate the consistency by predicting the brain activity of one held-out subject from all other subjects using 10 cross-validation splits for all subjects. However, for \datasetname{Tuckute2024} we use the theoretical estimate provided by \cite{tuckute2024driving}.
To aggregate across metrics, the model score (Pearson $r$ for Linear Predictivity) on each benchmark is normalized with the cross-subject consistency estimate ($\textnormal{normalized score} = \frac{\operatorname{max}(0,\textnormal{raw score})}{\textnormal{consistency}}$) and we report the final score for each model as the average across all considered benchmarks. Otherwise, the raw brain alignment refers to the average Pearson $r$ across datasets without normalization.

\section{Brain-Like Representations with Shallow Untrained Multihead Attention}
\label{sec:brain-like-repr}

We here investigate which architectural components drive the alignment of (untrained) transformer-based models to the human language system, and synthesize our findings into a simple untrained architecture. 

\begingroup
\renewcommand{\tabcolsep}{1.5pt}
\begin{table}[]
    \centering
    \caption{
    \textbf{Efficient Brain Alignment} on the five datasets for \ourmodel and different architectures, pretrained/untrained, with the number of FLOPs as a proxy for simplicity and efficiency. Each untrained model was evaluated using 5 random initializations, we here report the average. Fed2016 refers to the \datasetname{Fedorenko2016} dataset. }
    \begin{tabular}{lcccccc}
    \toprule
    \textbf{Model} (\textbf{MFLOPs}) & \textbf{Pereira2018} & \textbf{Blank2014} & \textbf{Fed2016} & \textbf{Tuckute2024} & \textbf{Wehbe2014} & \textbf{Average} \\
    \midrule
    \textbf{GPT2-Small} (170) & 0.38/0.16 & 0.10/0.05 & 0.27/0.27 & 0.29/0.21 & 0.11/0.05 & 0.23/0.15\\
    \textbf{GPT2-Med} (604) & 0.38/0.16 & 0.10/0.04 & 0.29/0.26 & 0.37/0.19 & 0.11/0.05 & 0.25/0.14\\
    \textbf{GPT2-Large} (1,420) & 0.39/0.16 & 0.09/0.05 & 0.30/0.25 & 0.32/0.21 & 0.08/0.04 & 0.23/0.14\\
    \textbf{GPT2-XL} (2,950) & 0.34/0.15 & 0.04/0.04 & 0.27/0.25 & 0.34/0.23 & 0.04/0.04 & 0.21/0.15\\
    \textbf{LLaMA-2-7B} (12,950) & 0.32/0.32 & 0.01/0.24 & 0.22/0.34 & 0.34/0.13 & 0.02/0.15 & 0.18/0.24\\
    \textbf{LLaMA-2-13B} (25,380) & 0.41/0.28 & 0.04/0.14 & 0.26/0.32 & 0.34/0.17 & 0.06/0.09 & 0.22/0.20\\
    \midrule
    \textbf{SUMA} (268) & - / 0.43 & - / 0.44 & - / 0.34 & - / 0.19 & - / 0.21 & - / \textbf{0.32}\\
    \bottomrule
    \end{tabular}
    \label{tab:brain-alignment}
\end{table}
\endgroup

\subsection{Isolating Critical Components of the Transformer Architecture}

We conduct a comprehensive ablation study of the architectural components in a single untrained \modelname{LLaMA} Transformer block to identify the key elements driving the alignment of (untrained) models with the HLS. Figure \ref{fig:ablations}(a) illustrates a Transformer block with the different components which we evaluate on its brain alignment in Figure \ref{fig:ablations}(b-d). 

\paragraph{Token Aggregation via Multihead Attention} 
The brain alignment of the last token vector alone, even when coupled with an MLP, is close to zero (Figure \ref{fig:ablations}(b), first two bars).
Aggregating tokens through an attention mechanism significantly increases the brain alignment (third bar), with higher alignment than even trained models (Table \ref{tab:brain-alignment}). 
Other architectural changes such as adding positional encoding or adding an MLP to the attention do not improve the brain alignment over the simple \code{Token-LayerNorm-Attention} model. We further find that encoding token frequency using a BPE tokenizer results in higher brain alignment than a simple word-based tokenization scheme.

We hypothesized that the aggregation of token vectors is a critical mechanism underlying high alignment of models.
Indeed we find that the simple mixing of tokens via a mean operation achieves brain alignment close to that of the attention architecture (gray bar). Increased diverse aggregation via more attention heads, i.e. multiple association mappings between different tokens in the context, improves brain alignment considerably (Figure \ref{fig:ablations}(c)). 




\paragraph{Recurrent Application of Shared Weights}
Since increased token mixing via multiple attention heads improved brain alignment, we explored whether further increasing mixing via the repeated application of the attention mechanism would yield further improvements. In neuroscience terms, this is typically referred to as \emph{recurrence}, wherein representations are processed multiple times through the same neurons. In ML terms, we unroll the same set of modules multiple times with \emph{shared weights}. This approach increases the computational depth while maintaining the same number of parameters, in similar spirit to \citep{dehghani2018universal, Lan2019ALBERTAL}.
In Figure \ref{fig:ablations}(d) we tested different numbers of unrolling steps and found that passing the output hidden states of the first pass to the same module again (i.e., two passes) leads to the best brain alignment.

\paragraph{\ourmodel} To summarize, we identified token aggregation via repeated multihead attention as key architectural components that drive the high brain alignment of untrained models. We combine these building blocks into an architecture that we term \ourmodel: tokens are processed through \code{LayerNorm} and \code{Multihead Attention} with two layers with shared weights (a simple form of recurrence). This model is much simpler than most other architectures, both in terms of its components but also in terms of its (untrained) parameter count. We quantify this simplicity as the number of floating point operations (FLOPs) and observe that \ourmodel{} provides highly efficient brain alignment (Table \ref{tab:brain-alignment}). As a sanity check that \ourmodel{}'s brain alignment results are not a byproduct of spurious features, we tested the brain alignment for different control conditions (see Appendix \ref{app:control-conditions-metrics}). 



When developing our \ourmodel{} model, we used the aggregate across three datasets to guide our design decisions. To measure distributional robustness, we evaluate on two other datasets: the \datasetname{Wehbe2014} dataset where participants read a chapter from Harry Potter and the Sorcerer’s Stone, and on the distributionally-diverse \datasetname{Tuckute2024} dataset consisting of brain responses to 1000 linguistically diverse single sentences. Although \ourmodel{} still provides an efficient trade-off between FLOPs and brain alignment, we observe a drop in alignment (Table \ref{tab:brain-alignment}) on the \datasetname{Tuckute2024} dataset compared to pretrained models. However, \ourmodel achieves the highest alignment score on the other held-out \datasetname{Wehbe2014} dataset.


\begin{figure}
    \centering
    \includegraphics[width=1\linewidth]{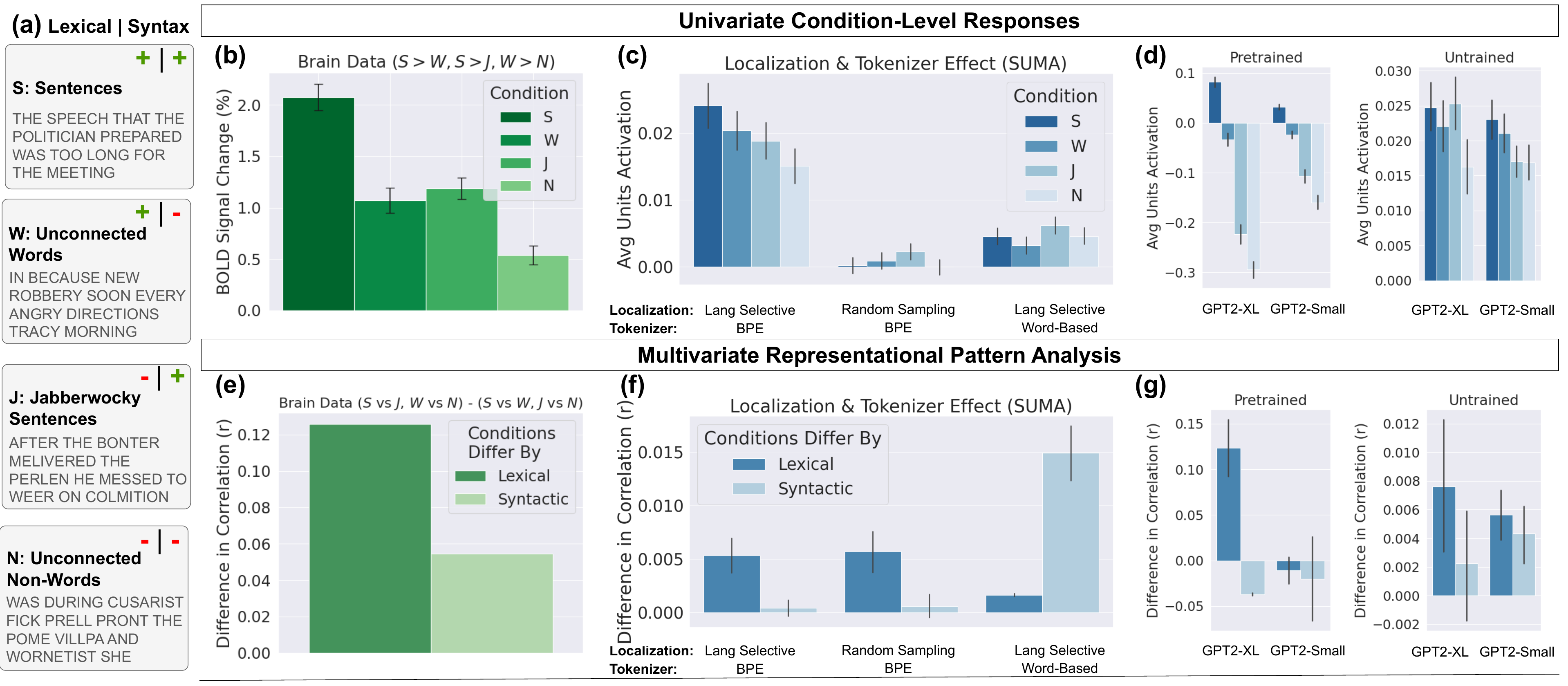}
    \caption{
        \textbf{Language Models Exhibit Similar Response Profiles as the HLS.} Brain ({\color{ggreen} green}) and model ({\color{blue} blue}) responses for Univariate Condition-Level Responses (Top Row) and Multivariate Representational Pattern Analysis (Bottom Row). Each untrained model plot is the average across 5 different random model initializations. The error bars are across the different initializations and conditions.
        \textbf{(a)} Examples of the four experimental conditions used in this analyses with the `+/-' signs denoting whether the condition contains lexical or syntactic information, respectively. 
        \textbf{(b)} Brain responses to the four conditions; data from \citep{Shain2023}. 
        \textbf{(c)} \ourmodel responses to the four conditions. Control experiments show the effect of unit localization (``Lang Selective'' vs ``Random Sampling'') and tokenization (``BPE'' vs ``Word-Based''). 
        \textbf{(d)} The same univariate analysis for \modelname{GPT2-XL} and \modelname{GPT2-Small} models. 
        \textbf{(e-g)} Same as (b-d) but for the multivariate analysis (Section \ref{sec:hls-response-profiles}). Brain data from extracted from reported results in \cite{Fedorenko2012}.
    }
    \label{fig:neural-analyses}
\end{figure}

\subsection{Language Models Exhibit Similar Response Profiles as the HLS}
\label{sec:hls-response-profiles}

To further validate \ourmodel's alignment with the human language system, we investigate whether it can replicate landmark neuroscience studies that qualitatively describe the response profiles in the HLS. 
Specifically, we closely follow the analyses in \cite{Fedorenko2010NewMF} and \cite{Fedorenko2012}, using the same set of experimental conditions which are widely used in neuroimaging studies examining lexical and syntactic processing. The four conditions are: 
(1) \code{Sentences}, denoted as \emph{S}, are well-formed sentences containing both lexical and syntactic information. 
(2) \code{Unconnected Words}, denoted as \emph{W}, are scrambled sentences containing lexical but not syntactic information. 
(3) \code{Jabberwocky Sentences}, denoted as \emph{J}, where content words are replaced by pronounceable non-words (such as ``pront'', or ``blay''), thus containing syntactic but not lexical information. 
(4) \code{Unconnected Non-Words}, denoted as \emph{N}, which are scrambled Jabberwocky sentences containing neither lexical nor syntactic information. Examples of the four experimental conditions are shown in Figure \ref{fig:neural-analyses}(a). Note that we use a disjoint set of \code{Sentences} and \code{Non-Words}for the original functional localization (Section \ref{sec:localization}). 

\paragraph{Models Predict Univariate Condition-Level Responses}
In this analysis, we record the responses of the localized units from \ourmodel when presented with stimuli from each of the four experimental conditions. Figure \ref{fig:neural-analyses}(b) displays empirical fMRI responses from the HLS in green (data from \cite{Shain2023}). The brain's language regions are highly sensitive to linguistic structure: responses to \emph{S} are numerically higher than all other conditions, while responses to both \emph{W} and \emph{J} are higher than those to \emph{N}, which contains neither structure nor meaning. 
Our \ourmodel{} model replicates these results (Figure \ref{fig:neural-analyses}(c)). We find that localization and tokenization are critical model components: using a BPE tokenizer and recording responses from the localized units show a similar response profile with as the brain recordings. However, when we randomly sample the same number of units from the model, or use a word-based tokenization scheme the resulting profile is no longer consistent with brain data. 
We observe similar response profiles in other architectures such as \modelname{GPT2-XL} and \modelname{GPT2-Small} (Figure \ref{fig:neural-analyses}(d)). We explore the effect of the number of localized units on this analysis in Appendix \ref{app:neural-analyses}.

\paragraph{Models Predict Multivariate Representational Patterns}
We follow the Multivoxel Pattern Analysis \citep{Haxby2001} conducted in \cite{Fedorenko2012}, where the authors demonstrate that the brain's language regions can more reliably discriminate between conditions that differ along the lexical dimension (\emph{S} vs \emph{J} and \emph{W} vs \emph{N}) than those differing along the syntactic dimension (\emph{S} vs \emph{W} and \emph{J} vs \emph{N}). To illustrate this analysis, consider the \emph{S} vs \emph{J} case: Each condition is divided into two sets (i.e., ($S_1$, $S_2$) and ($J_1$, $J_2$)). We then calculate the Pearson correlation between the localized unit responses to $S_1$ and $S_2$ (within-correlation) and the correlation between responses to $S_1$ and $J_2$ and $S_2$ and $J_1$ (in-between correlation). The difference between the within-correlation and the in-between correlation (we take the average across both in-between comparisons following \citep{Fedorenko2012}) in this case reflects sensitivity to lexical information, as lexical information is present in $S$ and absent in $J$, while both contain syntactic information. Following \cite{Fedorenko2012}, all $r$ values are Fisher transformed. 
Figure \ref{fig:neural-analyses}(e) compares the difference in correlation between the pairs of conditions that differ by lexical and syntactic information, showing that language voxels in the brain are more sensitive to lexical information. Repeating the same analysis in models, we again find that our \ourmodel{} model replicates the neuroscience observations -- but only when using a BPE tokenizer (Figure \ref{fig:neural-analyses}(f)). We observe a similar pattern in other language models (Figure \ref{fig:neural-analyses}(g)) and report more results in Appendix \ref{app:neural-analyses}.




\section{Language Modeling with Untrained Representations}
\label{sec:lm-results}

After finding that the representations in certain untrained models are highly aligned to the HLS---both quantitatively and also qualitatively---we asked whether they could then also support downstream language production.
We appended a trainable decoder to several untrained models to test if there is a benefit to using the representations encoded by them.

\subsection{Localized Untrained Units Improve Language Modeling}

\paragraph{Language Modeling Setup}
We train 6 models using causal language modeling. Two of these models, \modelname{SUMA+1} and \modelname{SUMA+2}, consist of a frozen untrained \ourmodel that uses a single forward pass (see Figure \ref{fig:ablations}) with an additional one or two trainable \modelname{LLaMA} Transformer blocks, respectively. For each SUMA+N model, we test two variants: one that takes the localized units as input, and another that takes the output of the final \ourmodel layer as input. Additionally, we train two control models that directly take the untrained tokens as input. We refer to these models as \modelname{Baseline-333M} and \modelname{Baseline-536M}, and the numbers correspond to the number of trainable parameters in the decoder Transformer block(s) and the language model head. All models are trained on \datasetname{WikiText-103} \citep{merity2016pointer} for 5 epochs, with a batch-size of 128, a context-window size of 512 tokens, and an initial learning rate of $5e^{-3}$ when training a single block ($5e^{-4}$ when training two blocks) that anneals to zero throughout training. See Appendix \ref{app:training-setup} for more details about the training setup. We do not train the token embeddings for the 6 models to keep them comparable as otherwise we will have to perform localization at every optimization step since the token embeddings will keep changing.

\begingroup
\setlength{\tabcolsep}{5pt}
\begin{table}[t]
    \centering
    \caption{
        \textbf{Language Modeling Perplexity and Behavioral Alignment Results.}
        \code{PPL} denotes the \datasetname{WikiText-103} test perplexity results, \code{Behavior} denotes the alignment of model surprisal to human reading times in the \datasetname{Futrell2018} benchmark. \code{Params} denotes the number of trainable parameters. \modelname{GPT2} models use a different tokenizer, making results not directly comparable. The best result within each block is highlighted in \textbf{bold}.
    }
    \begin{tabular}{llrrrr}
    \toprule 
    \textbf{Model} & \textbf{Input Repr (Trained)} & \textbf{Params} & \textbf{MFLOPs ($\downarrow$)} & \textbf{PPL ($\downarrow$)} & \textbf{Behavior ($\uparrow$)} \\
    \midrule
    \textsc{Baseline-333M} & Embeddings (\xmark)  & 333M & 667 & 60.26 & 0.446 \\
    \textsc{SUMA+1} & Final Layer (\xmark) & 333M & 801 & 30.48 & \textbf{0.452} \\
    \textsc{SUMA+1} & Localized Units (\xmark) & 333M & 801 & \textbf{22.17} & 0.451 \\
    \midrule
    \textsc{Baseline-536M} & Embeddings (\xmark) & 536M & 1070 & 20.22 & 0.442 \\
    \textsc{SUMA+2} & Final Layer (\xmark) & 536M & 1210 & 20.81 & 0.449 \\
    \textsc{SUMA+2} & Localized Units (\xmark) & 536M & 1210 & \textbf{16.51} & \textbf{0.453} \\
    \midrule 
    \textsc{LLaMA-2-7B} & Embeddings (\cmark)  & 7B & 13220 & \textbf{12.33} & \textbf{0.243} \\
    \textsc{LLaMA-2-13B} & Embeddings (\cmark) & 13B & 25700 & 16.29 & 0.221 \\
    \midrule
    \midrule
    \textsc{GPT2-Small} & Embeddings (\cmark) & 117M & 247 & 37.50 & \textbf{0.367} \\
    \textsc{GPT2-Medium} & Embeddings (\cmark) & 345M & 707 & 26.37 & 0.350 \\
    \textsc{GPT2-Large} & Embeddings (\cmark) & 762M & 1550 & 22.05 & 0.331 \\
    \textsc{GPT2-XL} & Embeddings (\cmark) & 1.5B & 3110 & \textbf{17.48} & 0.318 \\
    
    \bottomrule
    \end{tabular}
    \label{tab:ppl-results}
\end{table}
\endgroup

\begin{figure}
    \centering
    \includegraphics[width=1\linewidth]{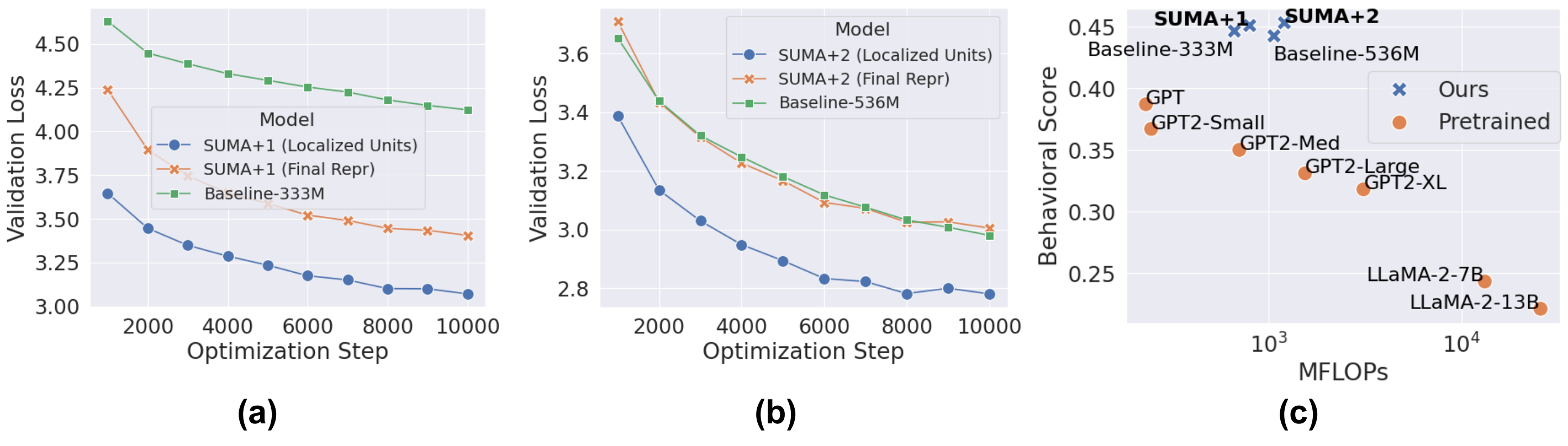}
    \caption{
        \textbf{Localized Untrained Units Provide Representations Suitable for Language Modeling.}
        \textbf{(a-b)} The \datasetname{WikiText-103} validation loss of \ourmodel variants and control models during training, when training (a) one Transformer block, and (b) two Transformer blocks. We train two variants of \ourmodel-based models: one that uses the output of the localized units as the input representation for the downstream decoder, and another that uses the final representation (Final Repr) of the untrained model as input. The baseline model refers to passing the static tokens directly to the decoder without any intermediate architecture. 
        \textbf{(c)} Behavioral alignment to human reading times in \datasetname{Futrell2018} as a function of model complexity and efficiency measured by the number of FLOPs. 
    }
    \label{fig:lm-results}
\end{figure}

\paragraph{Results}
Table \ref{tab:ppl-results} compares the \datasetname{WikiText-103} test perplexity and the behavioral score (described in the next section) of the different models considered in this work, along with various \modelname{GPT2} and \modelname{LLaMA-2} models for reference.\footnote{For \modelname{GPT2}, we use the results reported in \cite{gpt2}, which we closely reproduced, and for \modelname{LLaMA-2}, we evaluated the models ourselves using a context-window size of 512 tokens.} We report the floating point operations per second as a proxy for model complexity. 
We find that localized units in the untrained \ourmodel model provide effective representations for language modeling: Passing the randomly initialized tokens through \ourmodel and then extracting the localized units distributed across the network as input significantly improves perplexity and sample efficiency. This method performs better than taking the representation from the last layer or passing the tokens directly as input (Table \ref{tab:ppl-results}). This improvement is especially pronounced when appending a single trainable Transformer layer; increasing the number of parameters by adding another block reduces this gap (Figure \ref{fig:lm-results} (b)). 
In all cases, localizing units in \ourmodel as input not only achieves better final perplexity than the control models with similar number of parameters, but already reaches a lower validation loss early during training.

\subsection{State-of-the-art Human Reading Times Prediction}
Training decoder layers on top of our untrained models allows us to evaluate how similar model logit surprisal ratings are to human behavior. This enables us to test whether the entire proposed architecture (Figure \ref{fig:methods}) is reasonably consistent with the HLS's outputs beyond its internal representations.

\paragraph{Behavioral Benchmark}
We use the \datasetname{Futrell2018} benchmark, which has been widely used in previous research to measure linguistic behavior \citep{futrell_natural_2018, schrimpf-pnas, aw2023instructiontuning}. This dataset consists of self-paced reading times for naturalistic story materials from 180 participants. Per-word reading times provide a measure of incremental comprehension difficulty, a cornerstone of psycholinguistic research for testing theories of sentence comprehension \citep{Gibson1998, Smith2013}. We measure alignment by calculating the Pearson correlation between a model's cross-entropy loss for a specific token in the sequence and the average human per-word reading time. The loss for words that comprise of multiple tokens are added together before computing the correlation. 

\paragraph{Results}
Figure \ref{fig:lm-results}(c) shows the behavioral score as a function of the model's complexity (exact numbers in Table \ref{tab:ppl-results}). On reference \modelname{GPT2} and \modelname{LLaMA-2} models, we surprisingly observe that increasing the complexity reduces the behavioral score (see also \cite{oh2023does,shain2024large}.
Our shallow models all surpass the pretrained models by a large margin, with the most complex \modelname{SUMA+2} achieving a new state-of-the-art on this human reading times benchmark. 

\section{Discussion}
\label{sec:discussion}

\paragraph{Tokenization \& Multihead Attention}
Our findings identify token frequency and aggregation via multihead attention or a simple mean operation as critical components driving brain alignment (Section \ref{sec:brain-like-repr}). Utilizing a Byte Pair Encoding (BPE) tokenizer enhances brain alignment by effectively encoding token frequency, breaking down infrequent words into more tokens \citep{Kida1999BytePE, sennrich-etal-2016-neural}. This frequency-based tokenization scheme allow the model to more accurately capture linguistic nuances, aligning well with the sensitivity of humans to lexical frequency in language comprehension \citep{rayner1986lexical, Singh2016QuantifyingSC}. Further aggregating these tokens through a multihead attention mechanism then constructs diverse representations between each token pair within the context. The improved alignment from an increased number of attention heads is most likely due to greater expressivity achieved by multiple heads, enabling the model to presumably capture an array of context-dependent relations.

\paragraph{Encoding via Architectural Priors Aid Language Decoding}
We found that passing localized units as token representations to downstream trainable modules leads to significantly better language modeling performance (Section \ref{sec:lm-results}). Through structural priors inherent in the architecture, the representations are potentially constrained to a space from which it is easier to generalize. In neuroscience terms, this approach follows an encoder-decoder setup where the untrained \ourmodel{} encoder provides generic representations that the downstream trainable decoder can then make effective use of. Interestingly, our combined models are significantly more aligned to human reading times than even much larger pretrained models. Similar observations have been reported by \citet{steuer2023large}, \citet{oh2023does}, and \citet{shain2024large}. 

\paragraph{A Need for Better Brain Benchmarks}
Although we are excited about the strong brain alignment of our model, we do not believe it to be a perfect model of the human language system. 
First, while cross-subject consistency estimates typically attempt to compute an ``upper bound'' or ceiling of how well the best possible model could perform, this estimate is routinely surpassed by the best models (Table \ref{tab:brain-alignment}). We believe the data consistency is under-estimated by trying to predict noisy data (target subject) from other noisy data (held-out subject pool; whereas model predictors are not noisy). We hope future brain recordings of language will include more stimulus repetitions to increase this signal-to-noise (SNR) ratio and allow more accurate ceiling estimates.
Second, we observed inconsistencies between different datasets (Table \ref{tab:brain-alignment}) and especially different metrics (Appendix \ref{app:add-metrics}). 
Third, the \datasetname{Fedorenko2016} dataset show high alignment even when a random sequence of tokens of the same length is passed to untrained models. This appears to be because most stimuli in this dataset consist of unique words, making them indistinguishable from the original sentence in untrained models. Highlighting the importance of linguistic diversity in the stimuli set presented to participants in neuroimaging studies.
Finally, it is important to note that our final model architecture is based on the aggregate score across multiple datasets, so it may not represent the best possible alignment achievable with a shallow untrained model on individual datasets.
Despite a (perpetual) wish for data with higher SNR, we find that simple models such as single-token only or the sentence length and position do \emph{not} necessarily yield high alignment scores, making us confident in \modelname{}'s utility.


\section{Conclusion}

By identifying the key components that contribute to neural alignment between language models and the human language system, we demonstrate that a shallow untrained multihead attention network can achieve significant alignment with the language regions in the brain. We first perform functional localization to identify language-selective units within a language model, analogous to how neuroscientists localize the language system in the human brain, and benchmark model-to-brain alignment across five different brain recordings. We identify token aggregation via repeated application of the multihead attention mechanism combined with the tokenization strategy that implicitly encodes token frequency, enhance the expressivity and effectiveness of the model. We further demonstrate that those localized units exhibit similar response profiles as shown in landmark studies in language neuroscience \citep{Fedorenko2024}. Our findings suggest that the human language system might be simpler than previously thought, potentially following a hierarchical structure similar to the processing of vision \citep{dicarlo2012} and speech \citep{Gwilliams2024Speech}. This conceptualization aligns with our model's ability to support downstream language production by acting as a feature encoder that provides suitable representations for a downstream decoder. Furthermore, our work underscores the need for datasets with higher signal-to-noise-ratio, and consistent metrics to accurately benchmark and evaluate model-to-brain alignment (see Appendix \ref{app:add-metrics}). Together, these efforts offer a new perspective on LLMs in the context of brain alignment, emphasizing that achieving high alignment does not necessarily require large-scale or highly complex models.

\begin{ack}
We would like to thank Abdülkadir Gökce, Yingtian Tang, Ben Lonnqvist, Hannes Mehrer, Muhammad ElNokrashy, Eghbal Hosseni, Anna Ivanova, and Ev Fedorenko for their valuable feedback and insightful discussions. We also thank Syrielle Montariol for her feedback on the final manuscript.
\end{ack}

\bibliography{references}
\bibliographystyle{apalike}

\newpage

\appendix

\section*{Appendix}

\section{Neuroimaging \& Behavioral Datasets}
\label{app:datasets}

\subsection{Neuroimaging Datasets}
\label{app:neural-datasets}

\paragraph{\cite{pereira_toward_2018}} This dataset consists of fMRI activations (blood-oxygen-level-dependent; BOLD responses) recorded as participants read short passages presented one sentence at a time for 4 s. The dataset is composed of two distinct experiments: one with 9 subjects presented with 384 sentences, and another with 6 subjects presented with 243 sentences each. The passages in each experiment spanned 24 different topics. The results reported for this dataset are the average alignment across both experiments after normalizing with their respective cross-subject consistency estimates.

\paragraph{\cite{blank_functional_2014}} This dataset also involves fMRI signals but recorded from only 12 functional regions of interest (fROI) instead of the higher resolution signal used by \citet{pereira_toward_2018}. The data was collected from 5 participants as they listened to 8 long naturalistic stories that were adapted from existing fairy tales and short stories \citep{futrell_natural_2018}. Each story was approximately 5 minutes long, averaging up to 165 sentences, providing a much longer context length than the other neuroimaging datasets except \datasetname{Wehbe2014}. When measuring brain alignment we use the input stimuli of the last 32 TRs as the model's context.

\paragraph{\cite{fedorenko2016}} This dataset captures ECoG signals from 5 participants as they read 8-word-long sentences presented one word at a time for 450 or 700 ms. Following \cite{schrimpf-pnas} we select the 52/80 sentences that were presented to all participants. 

\paragraph{\cite{tuckute2024driving}} 
This dataset is used to measure distributional robustness of our model, and is tested only after confirming all our design choices based on the preceding datasets. In this dataset, 5 participants read 1000 6-word sentences presented on one sentence at a time for 2 s. BOLD responses from voxels in the language network were averaged within each participant and then across participants to yield an overall average language network response to each sentence. The stimuli used span a large part of the linguistic space, enabling model-brain comparisons across a wide range of single sentences. Sentence presentation order was randomized across participants. The averaging of sentences across participants effectively minimizes the effect of temporal autocorrelation/context in this dataset. In combination with the diversity in linguistic materials, this dataset presents a particularly challenging dataset for model evaluation.

\paragraph{\cite{Wehbe2014}}
We similarly evaluate on this dataset after confirming our design choices on the first three datasets. The fMRI data was collected from 8 native English speakers while they read Chapter 9 of ``Harry Potter and the Sorcerer’s Stone.'' Participants were familiar with the Harry Potter series, and were reminded of the events leading up to Chapter 9 before the experiment. They read the chapter using Rapid Serial Visual Presentation (RSVP), with each word displayed for 0.5 seconds, over 45 minutes, covering 5,000 words. Since functional localization was not performed by the authors, we conduct the the localization ourselves for measuring brain alignment by extracting the top-10\% voxels from each anatomically defined language region from both the left and right hemisphere using a probabilistic atlas of the human language system \citep{Lipkin2022}. We exclude the angular gyrus since it has been found to consistently respond differently than other regions in the language system despite other similarities such as responding more strongly to sentences than unconnected non-words \citep{Fedorenko2024}.


\subsection{Cross-Subject Consistency Estimates}
\label{app:crosssubject}

Table \ref{tab:ceilings} shows the cross-subject consistency estimates (known as noise-ceiling in previous work) that we compute to normalize the raw correlation before aggregating the results across datasets and metrics. The method in which we compute those estimates is described in the main paper for the linear predictivity metric in Section \ref{sec:benchmarks} and for the non-parameteric metrics in Appendix \ref{app:add-metrics-details}. The \datasetname{Tuckute2024} dataset can only work with a regression-based metric since it contains only one value for each stimulus that we attempt to predict.

\begingroup
\setlength{\tabcolsep}{2pt}
\begin{table}[t]
    \centering
    \caption{\textbf{Cross-Subject Consistency Estimates} used in this work for the three metrics considered. \datasetname{Per2018} is the \datasetname{Pereira2018} dataset which consists of two separate experiments \{384, 243\} as described in Appendix \ref{app:neural-datasets} each with its own estimate. \datasetname{Fed2016} is the \datasetname{Fedorenko2016} dataset.}
    \begin{tabular}{lcccccc}
    \toprule 
     & \textbf{\datasetname{Per2018.384}} & \textbf{\datasetname{Per2018.243}} & \textbf{\datasetname{Fed2016}} & \textbf{\datasetname{Blank2014}} & \textbf{\datasetname{Tuckute2024}} & \textbf{\datasetname{Wehbe2014}} \\
    \midrule
    \textbf{Linear} & 0.3430 & 0.2826 & 0.2013 & 0.1621 & 0.5592 & 0.2134 \\
    \textbf{CKA} & 0.1247 & 0.1128 & 0.1547 & 0.0319 & - & - \\
    \textbf{RDM} & 0.0808 & 0.0802 & 0.0894 & 0.0201 & - & - \\
    \bottomrule
    \end{tabular}
    \label{tab:ceilings}
\end{table}
\endgroup

\subsection{Behavioral Datasets}

\paragraph{\cite{futrell_natural_2018}}
This dataset consists of self-paced reading times for each word from 180 participants. The stimuli include 10 stories from the Natural Stories Corpus \citep{futrell_natural_2018}, similar to \datasetname{Blank2014}. Each participant read between 5 and all 10 stories.  

\section{Brain Alignment Using Additional Metrics}
\label{app:add-metrics}

\subsection{Metrics: Centered Kernel Alignment \& Representational Dissimilarity Matrices}
\label{app:add-metrics-details}

We report here the results using additional metrics. Specifically, we use Centered Kernel Alignment \citep{cka} and Representational Dissimilarity Matrices \citep{rdm}. Unlike training a linear regression, both of those metrics are non-parameteric and therefore do not require any additional training. However, they exhibit unusual behavior when presenting the model with control stimuli, such as random input, as shown in Figure \ref{fig:suma-baselines-metrics} and discussed in Section \ref{app:control-conditions-metrics}, which is the reason why we opted for Linear Predictivity as our primary metric.

\paragraph{Centered Kernel Alignment (CKA)} \citet{cka} introduced CKA as a substitute for Canonical Correlation Analysis (CCA) to assess the similarity between neural network representations. Unlike linear predictivity, it is a non-parameteric metric and therefore does not require any additional training. CKA is particularly effective with high-dimensional representations, and its reliability in identifying correspondences between representations in networks trained from different initializations \citep{cka}.
  
\paragraph{Representational Dissimilarity Matrices (RDM)} 
\cite{rdm} introduced RDMs as a solution to the challenge of integrating brain-activity measurements, behavioral observations, and computational models in systems neuroscience. RDMs are part of a broader analytical framework referred to as representational similarity analysis (RSA). In practical terms, to compute the dissimilarity matrix for an $N$-dimensional network's responses to $M$ different stimuli, an $M$×$M$ matrix of distances between all pairs of evoked responses is generated for both brain activity and the language model's activations \cite{harvey2023duality}. The correlation between these two matrices is then used as a measure of brain alignment.

\paragraph{Estimating Cross-Subject Consistency}
For both CKA and RDM, we compute the similarity between two halves of subjects over all possible combinations, instead of predicting brain activity of one subject using all other subjects as we did for Linear Predictivity. This approach yields a higher estimate, which we believe more accurately reflects the true performance ceilings, as many of the models we tested exceeded the previously estimated ceiling.

\subsection{Results: Isolating Critical Components of the Transformer Architecture}
We here report the same results as in Figure \ref{fig:ablations} but as aggregate on the three validation datasets and the three metrics. The aggregate result is averaged across the 9 benchmarks after being normalized relative to each cross-subject consistency estimate. See Figure \ref{fig:trained-untrained-metrics} for more results on untrained models and their trained counterparts.

\begin{figure}
    \centering
    \includegraphics[width=1\linewidth]{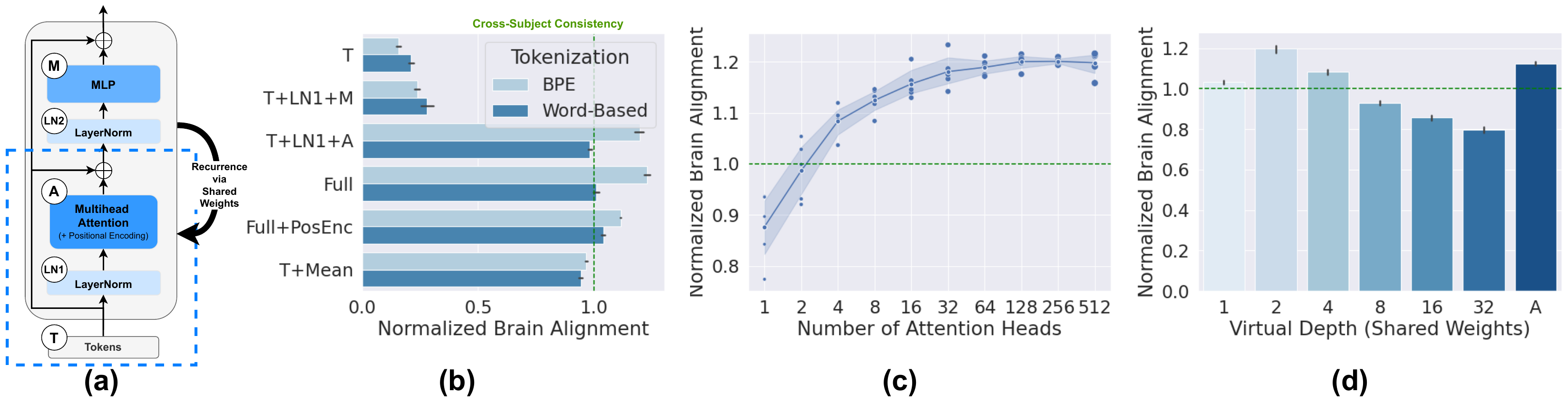}
    \caption{
    \textbf{Isolating Critical Components of the Transformer Architecture.}
    Similar plots as Figure \ref{fig:ablations} but aggregating the results across the three metrics on the three validation datasets instead of only using Linear Predictivity. Each experiment is repeated 5 across different random initializations.
    \textbf{(a)} Transformer block with components labeled for the ablation study in (b). The {\color{blue} blue} dashed box indicates our final model \ourmodel.
    \textbf{(b)} Brain alignment of a two-layer \ourmodel with shared weights on different architectural ablations and 512 attention heads. We use the model representation of the last token, except for \code{T+Mean}. Labels refer to (a).
    \textbf{(c)} Increasing the number of attention heads increases brain alignment. We use here the \code{T+LN+A} architecture with a depth of two.
    \textbf{(d)} Brain alignment of the \code{T+LN+A} model as a function of the number of unrolling steps. Adaptive refers to adaptive depth relative to the number of tokens.
    }
    \label{fig:ablations-metrics}
\end{figure}

\paragraph{Architectural Ablation}
Figure \ref{fig:ablations-metrics}(b) shows those results in the architectural ablation study. We can see the same effect shown before that highlights the importance of token aggregation either through taking a simple mean or via the multihead attention mechanism. However, in the case when using the three metrics, the best model is when using the full Transformer block architecture without positional encoding, although by just a small margin compared to the \code{T+LN+A}, which we used as the \ourmodel architecture.

\paragraph{Multihead Attention}
Figure \ref{fig:ablations-metrics}(c) shows a similar to the conclusions we reached with the linear metric, we here see the same effect that increasing the number of attention heads increases brain alignment, reaching a plateau at around 512 attention heads for a dimensionality of 4096. This is using the \ourmodel architecture described in the main paper.

\paragraph{Increasing Depth via Shared Weights}
Figure \ref{fig:ablations-metrics}(d) shows a similar trend as well for the effect of unrolling the \ourmodel model $L$ times in the depth dimension. Interestingly increasing the effective depth to 2 also results in the best alignment for those datasets.

\subsection{Results: Brain Alignment Controls on Each Metric}
\label{app:control-conditions-metrics}

\begin{figure}
    \centering
    \includegraphics[width=1\linewidth]{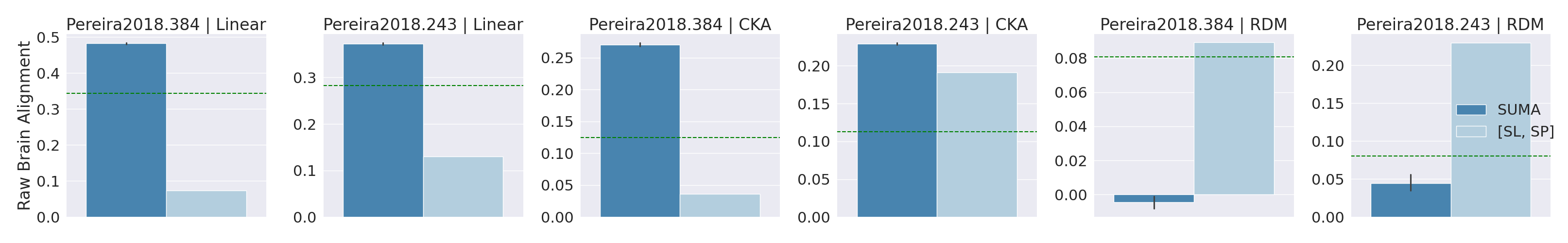}
    \caption{
        \textbf{Sentence Length and Position Does Not Explain All Variance Predicted By \ourmodel} We here replicate the experiment \citet{feghhi2024large} that claims \emph{sentence length} (SL) and \emph{sentence position} (SP) can account for all variance expalined by untrained models on \datasetname{Pereira2018}, which is not the case with our method.
    }
    \label{fig:sent-len-pos-baseline}
\end{figure}

\begin{figure}[h]
    \centering
    \includegraphics[width=1\linewidth]{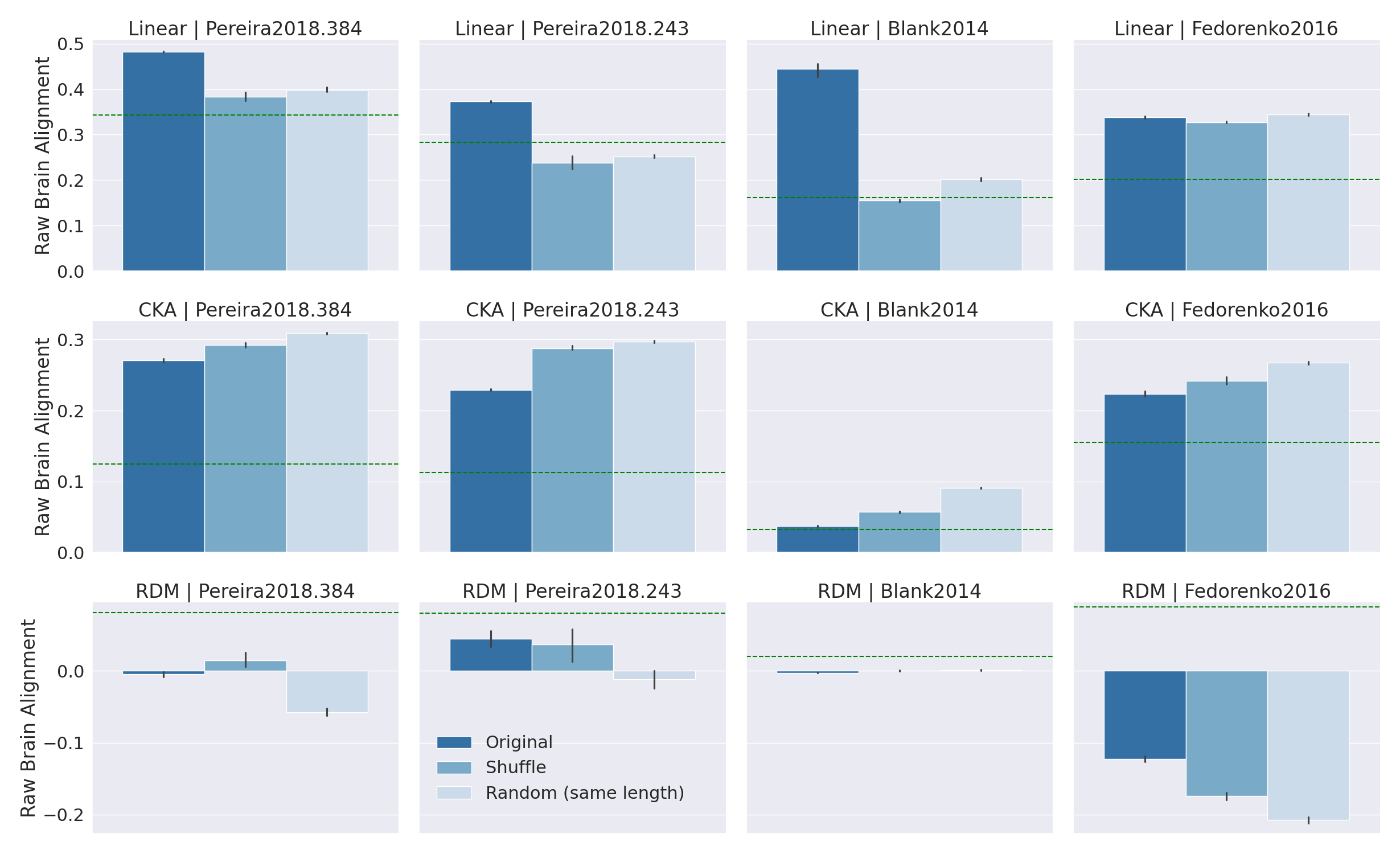}
    \caption{\textbf{\ourmodel Brain Alignment with Respect to Control Conditions:} The raw alignment (Pearson correlation scores) for three metrics and three datasets are reported separately based on the input condition. Each plot includes bars in the following order: \textbf{Original} indicates that the sentence is passed correctly without modifications. \textbf{Shuffle} means the tokens in each sequence are shuffled. \textbf{Random (same length)} means uniformly sampling the same number of tokens from the entire vocabulary and passing them to the model. Results from \datasetname{Pereira2018} are divided into its two respective experiments. Error bars show the variance across five different random seeds. }
    \label{fig:suma-baselines-metrics}
\end{figure}

\begin{figure}
    \centering
    \includegraphics[width=1\linewidth]{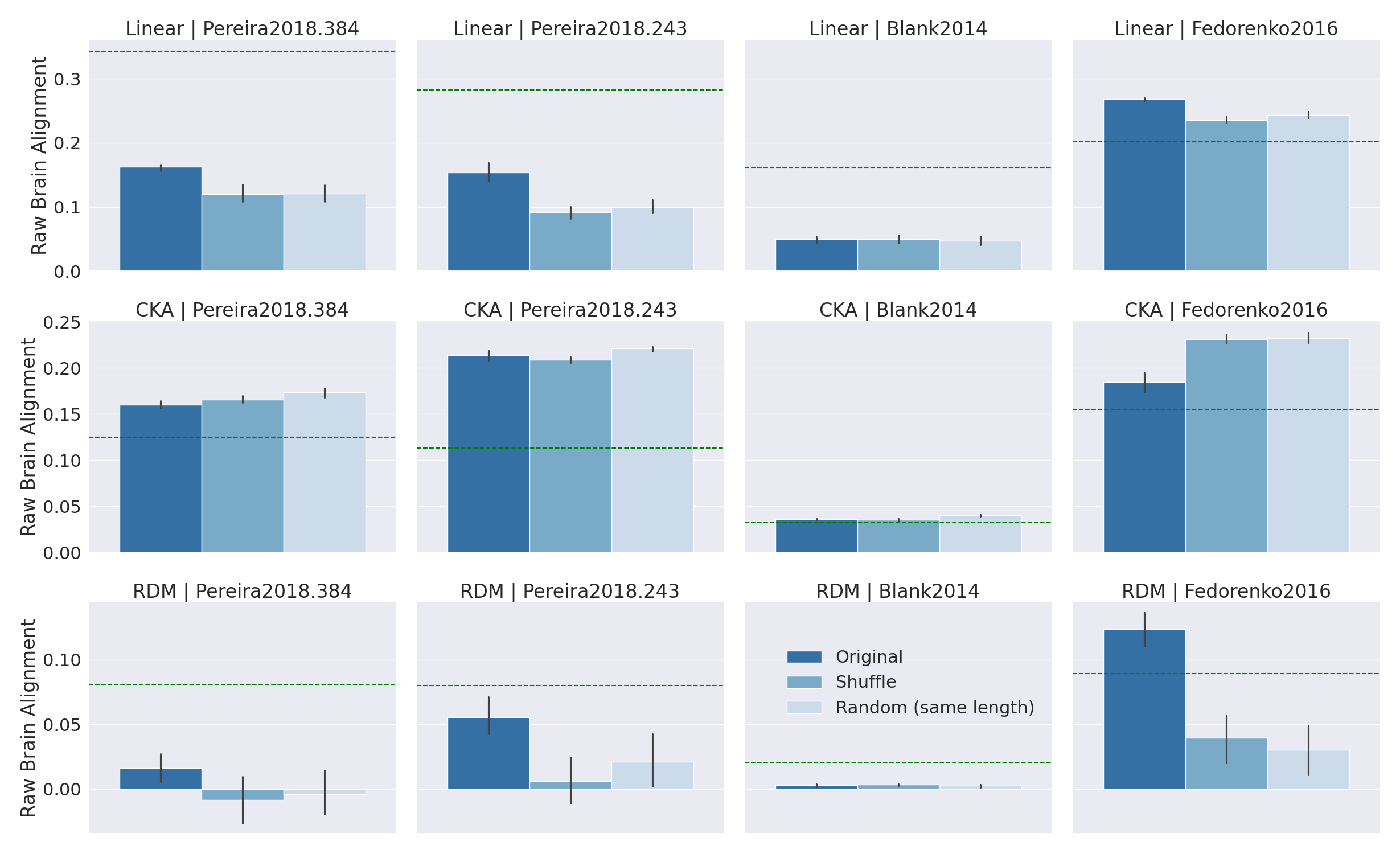}
    \caption{\textbf{Untrained GPT2 Brain Alignment with Respect to Control Conditions:} The raw alignment (Pearson correlation scores) for three metrics and three datasets are reported separately based on the input condition. Each plot includes bars in the following order: \textbf{Original} indicates that the sentence is passed correctly without modifications. \textbf{Shuffle} means the tokens in each sequence are shuffled. \textbf{Random (same length)} means uniformly sampling the same number of tokens from the entire vocabulary and passing them to the model. Results from \datasetname{Pereira2018} are divided into its two respective experiments. Error bars show the variance across five different random seeds.}
    \label{fig:gpt2-baselines-metrics}
\end{figure}

\begin{figure}
    \centering
    \includegraphics[width=1\linewidth]{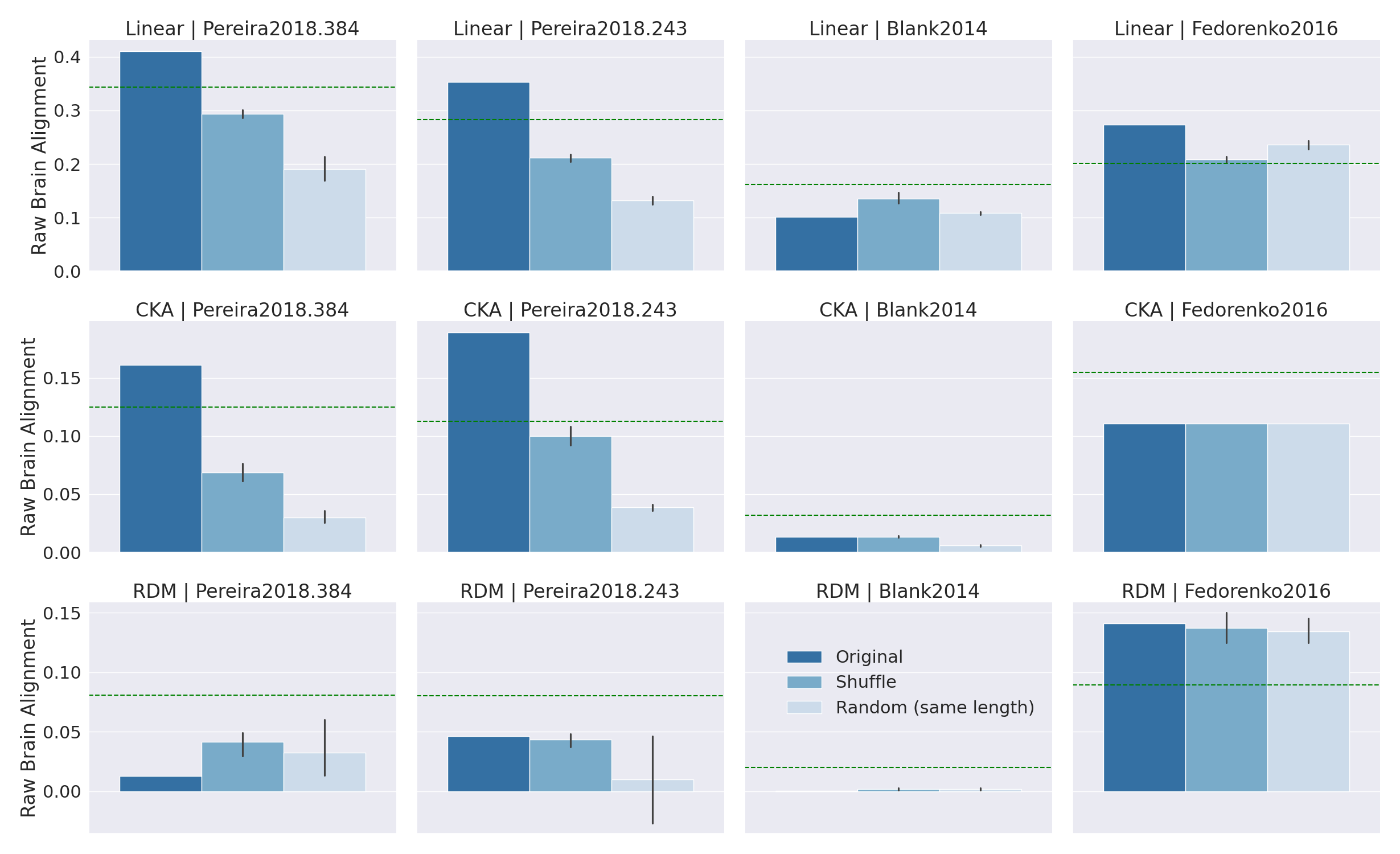}
    \caption{\textbf{Pretrained GPT2 Brain Alignment with Respect to Control Conditions:} The raw alignment (Pearson correlation scores) for three metrics and three datasets are reported separately based on the input condition similar to Figure \ref{fig:gpt2-baselines-metrics}.}
    \label{fig:gpt2-pretrained-baselines-metrics}
\end{figure}

We opted for Linear Predictivity as our primary metric since CKA and RDM exhibited unexpected behavior when given certain control sentences as input. Specifically, we compare the raw brain alignment of the original sentence under a given metric to the following conditions: to control for word order, we shuffle the tokens randomly at each point we measure alignment; to control for sequence length, we randomly sample the same number of tokens as in the original context from the model's vocabulary. Figures \ref{fig:suma-baselines-metrics}, \ref{fig:gpt2-baselines-metrics} and \ref{fig:gpt2-pretrained-baselines-metrics} shows the raw brain alignment for each metric and dataset independently for \ourmodel, untrained \modelname{GPT2-Small} and pretrained \modelname{GPT2-Small} respectively. For \ourmodel, CKA significantly surpasses the cross-subject consistency estimate and shows an opposite trend to what is expected, with a completely random sequence of tokens yielding the highest alignment in some cases. RDM similarly exhibits unexpected responses on some datasets. Untrained \modelname{GPT2-Small} shows a similar trend.
However, for pretrained models like \modelname{GPT2-Small}, CKA behaves as expected except for \datasetname{Fedorenko2016}, while RDM continues showing unexpected trends for pretrained models as well. Linear Predictivity was the only metric that consistently behaved as expected across all models when presented with control conditions. Moreover, we replicated the same experiment reported in \cite{feghhi2024large} that claims sentence length and sentence position accounts for all variance predicted by untrained \modelname{GPT2-XL} for the \datasetname{Pereira2018} dataset, and find that this is not the case for \ourmodel (see Figure \ref{fig:sent-len-pos-baseline}).


\clearpage

\subsection{Results: Pretrained Models and Untrained Counterparts}
\begin{figure}[h]
    \centering
    \includegraphics[width=1\linewidth]{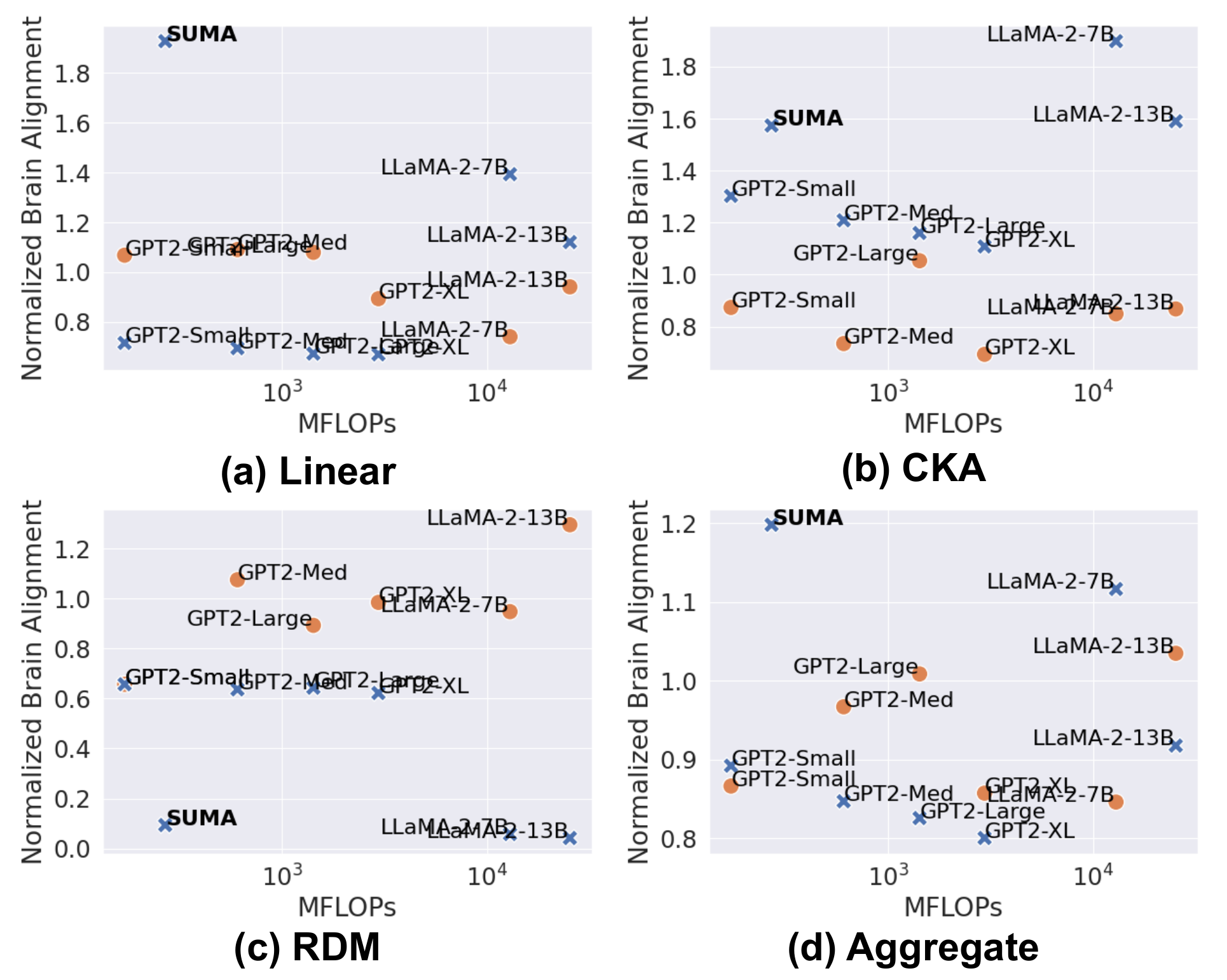}
    \caption{
    \textbf{Brain Alignment Results For Different Metrics} 
    Brain alignment for pretrained models and their untrained counteparts aggregated across the three validation datasets for:
    \textbf{(a)} Linear, \textbf{(b)} CKA, \textbf{(c)} RDM, and \textbf{(d)} Aggregate, with \ourmodel there as reference. MFLOPs---plotted in log-scale---is used as a proxy for the model's complexity.}
    \label{fig:trained-untrained-metrics}
\end{figure}

Figure \ref{fig:trained-untrained-metrics} shows the normalized brain alignment across the three metrics and three datasets for pretrained models and their untrained counterparts. The untrained models are evaluated 5 times with different initializations, and are initialized according to the default HuggingFace implementation \citep{Wolf2019HuggingFacesTS}. Interestingly, on those benchmarks, the untrained \modelname{LLaMA-2-7B} achieves higher alignment than its pretrained counterpart, while a model such as untrained \modelname{GPT2-XL} underperforms significantly compared to the pretrained \modelname{GPT2-XL}. 

\clearpage

\section{Univariate and Multivariate Representational Analyses}
\label{app:neural-analyses}

\begin{figure}
    \centering
    \includegraphics[width=1\linewidth]{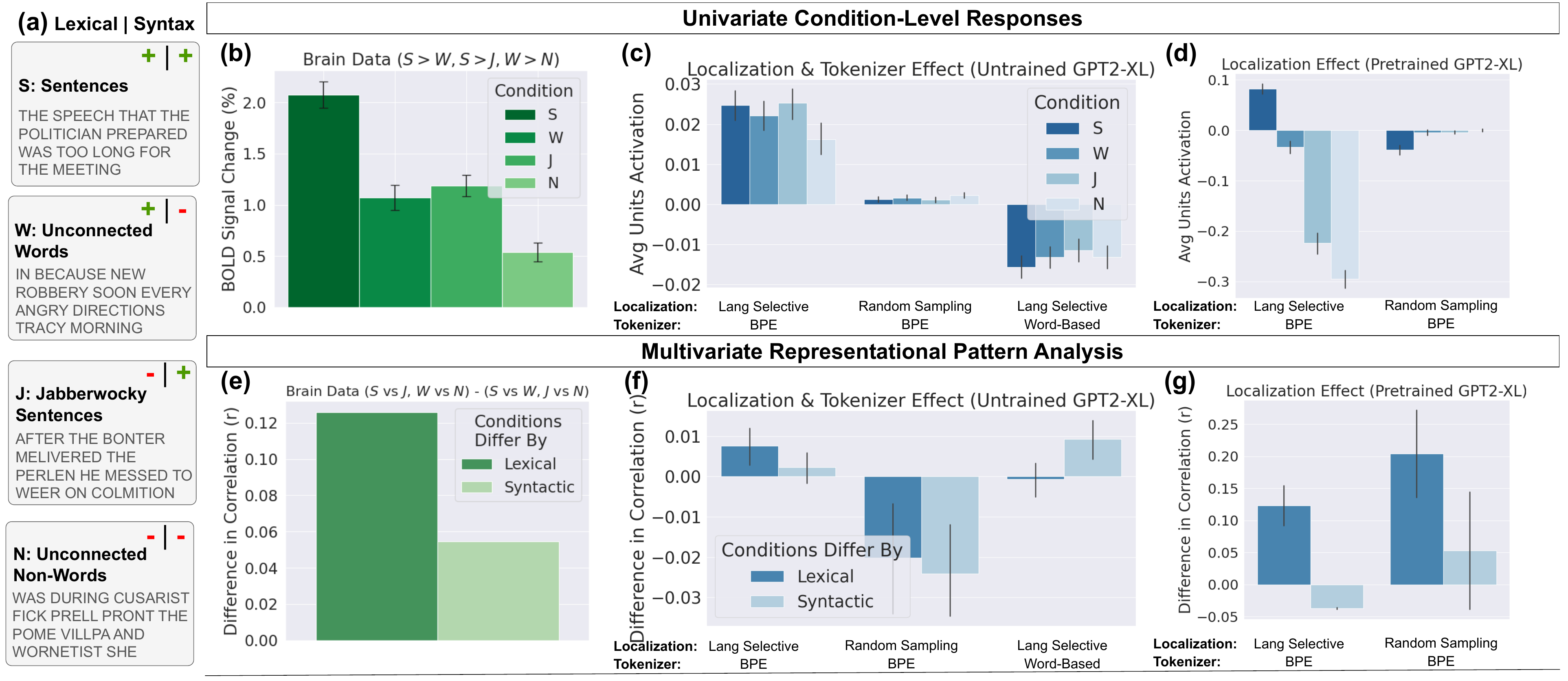}
    \caption{
        \textbf{Language Models Exhibit Similar Response Profiles as the HLS.} This is a similar to Figure \ref{fig:neural-analyses} but analyzes the localization and tokenization effect on untrained and pretrained \modelname{GPT2-XL}. Brain ({\color{ggreen} green}) and model ({\color{blue} blue}) responses for Univariate Condition-Level Responses (Top Row) and Multivariate Representational Pattern Analysis (Bottom Row). Each untrained model plot is the average across 5 different random model initializations. The error bars are across the different initializations and conditions.
        \textbf{(a)} Examples of the four experimental conditions used in this analyses with the `+/-' signs denoting whether the condition contains lexical or syntactic information, respectively. 
        \textbf{(b)} Brain responses to the four conditions; data from \citep{Shain2023}. 
        \textbf{(c)} Untrained GPT2-XL responses to the four conditions. Control experiments show the effect of unit localization (``Lang Selective'' vs ``Random Sampling'') and tokenization (``BPE'' vs ``Word-Based''). 
        \textbf{(d)} The same univariate analysis but for pretrained \modelname{GPT2-XL}. 
        \textbf{(e-g)} Same as (b-d) but for the multivariate analysis (Section \ref{sec:hls-response-profiles}). Brain data from extracted from reported results in \cite{Fedorenko2012}.
    }
    \label{fig:neural-analyses-gpt2xl}
\end{figure}

Figure \ref{fig:neural-analyses-gpt2xl} shows the same analyses described in Section \ref{sec:brain-like-repr} but focuses on responses from untrained and pretrained \modelname{GPT2-XL}. Similar to \ourmodel, Figure \ref{fig:neural-analyses-gpt2xl}(c) shows that the localized units in an untrained \modelname{GPT2-XL} exhibit a similar response profile to the human language system when presented with same set of conditions using a BPE tokenizer, whereas randomly sampling the same number of units or using a word-based tokenization strategy does not yield a similar profile. Figure \ref{fig:neural-analyses-gpt2xl}(d) shows the same univariate analysis but for pretrained \modelname{GPT2-XL}. Note that \modelname{GPT2-XL} was trained using BPE, therefore we do not evaluate the pretrained version on the word-based tokenization strategy.

Figure \ref{fig:neural-analyses-gpt2xl}(f) shows a similar pattern but for the multivariate representational analysis, where the localized units of untrained \modelname{GPT2-XL} is able to distinguish more reliably between inputs that differ along the lexical dimension than the syntactic dimension similar to what has been shown in neuroscience studies \citep{Fedorenko2012}. This is only true when using BPE and for the localized set of units. Interestingly, for pretrained \modelname{GPT2-XL} it does not seem to matter whether we localize or randomly sample the same number of units to exhibit a similar response profile to the brain.

\section{Effect of the Number of Localized Units}
\label{app:num-units}

\subsection{Effect on Brain Alignment}

\begin{figure}
    \centering
    \includegraphics[width=1\linewidth]{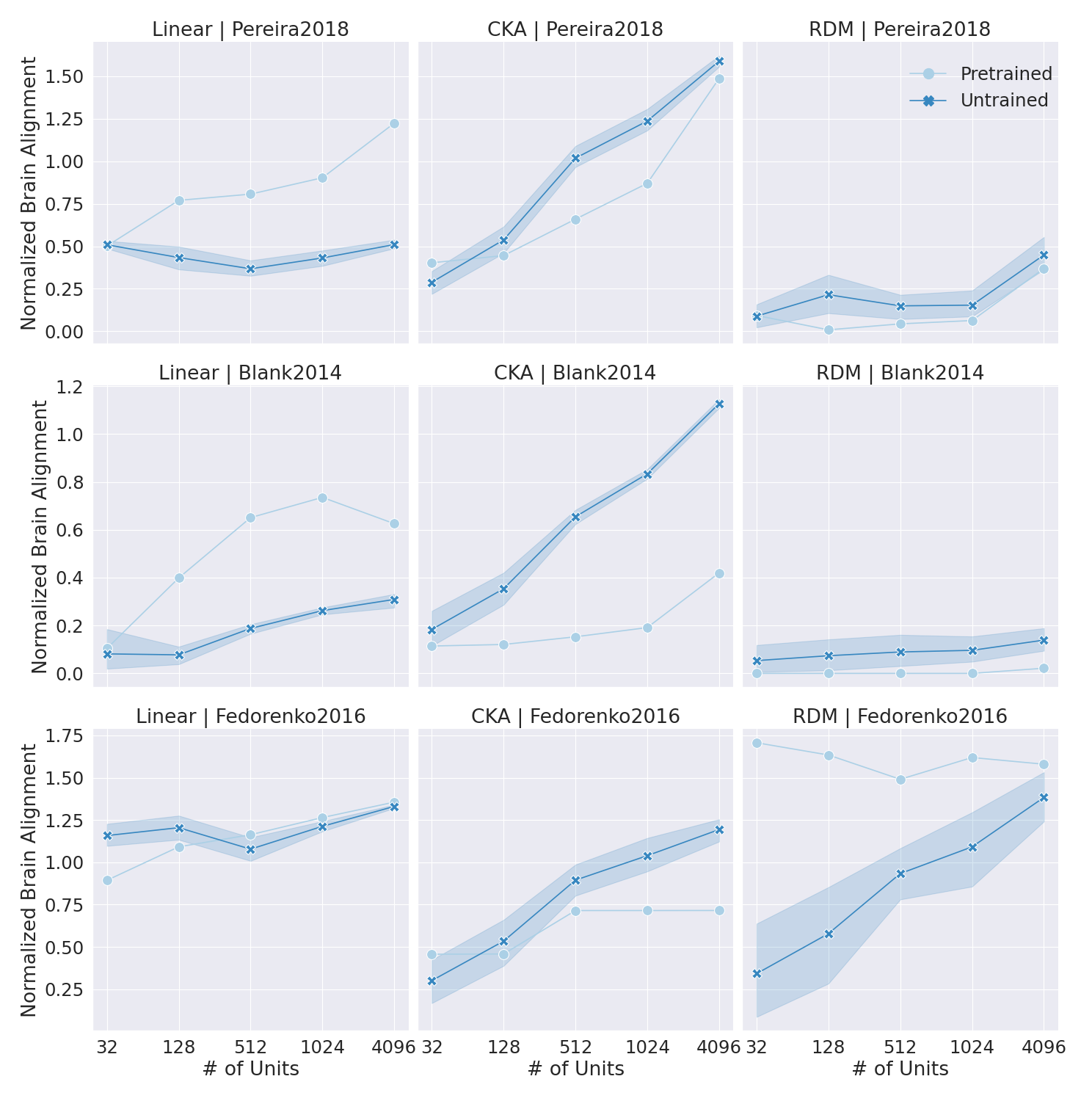}
    \caption{
        \textbf{Pretrained and Untrained GPT2-Small Brain Alignment As a Function of the Number of Localized Units}
        Here, we vary the number of top-k units localized and measure brain alignment across three validation datasets (rows) and three metrics (columns). The untrained model is evaluated with 5 different initializations for each $k$, and the average alignment is reported along with the variance, representing the 95\% confidence interval.
    }
    \label{fig:num-units-effect}
\end{figure}

Figure \ref{fig:num-units-effect} shows the effect of localizing the top-k units for different values of $k$. The results are presented for each metric and dataset separately using both pretrained and untrained \modelname{GPT2-Small}. The untrained result is the average across 5 different initializations for each $k$. The Linear Predictivity metric (first column) shows an interesting trend where the untrained models are more or less consistent as we increase the value of $k$ from 32 to 4096, whereas is the pretrained version increases more noticeably for the three datasets. The CKA metric (second column) on the other hand displays a consistent increase in alignment as we increase the number of localized units for both pretrained and untrained models demonstrating that dimensionality size does affect the raw correlation significantly. Finally, RDM (third column) shows a different trend for the \datasetname{Fedorenko2016} dataset where the untrained models only benefit from an increase in dimensionality size while pretrained does not, but for the other datasets the effect is much less pronounced. Note that when experimenting with linear regression without regularization the effect of the number of units was mich more significant with each benchmark (metric x dataset) combination behaving differently. Future work should focus on developing a method that selects the number of units based on each model's size for example and the dataset in question, since each dataset measures different number of voxels/regions.


\subsection{Effect on Univariate and Multivariate Representational Analyses}

\begin{figure}
    \centering
    \includegraphics[width=1\linewidth]{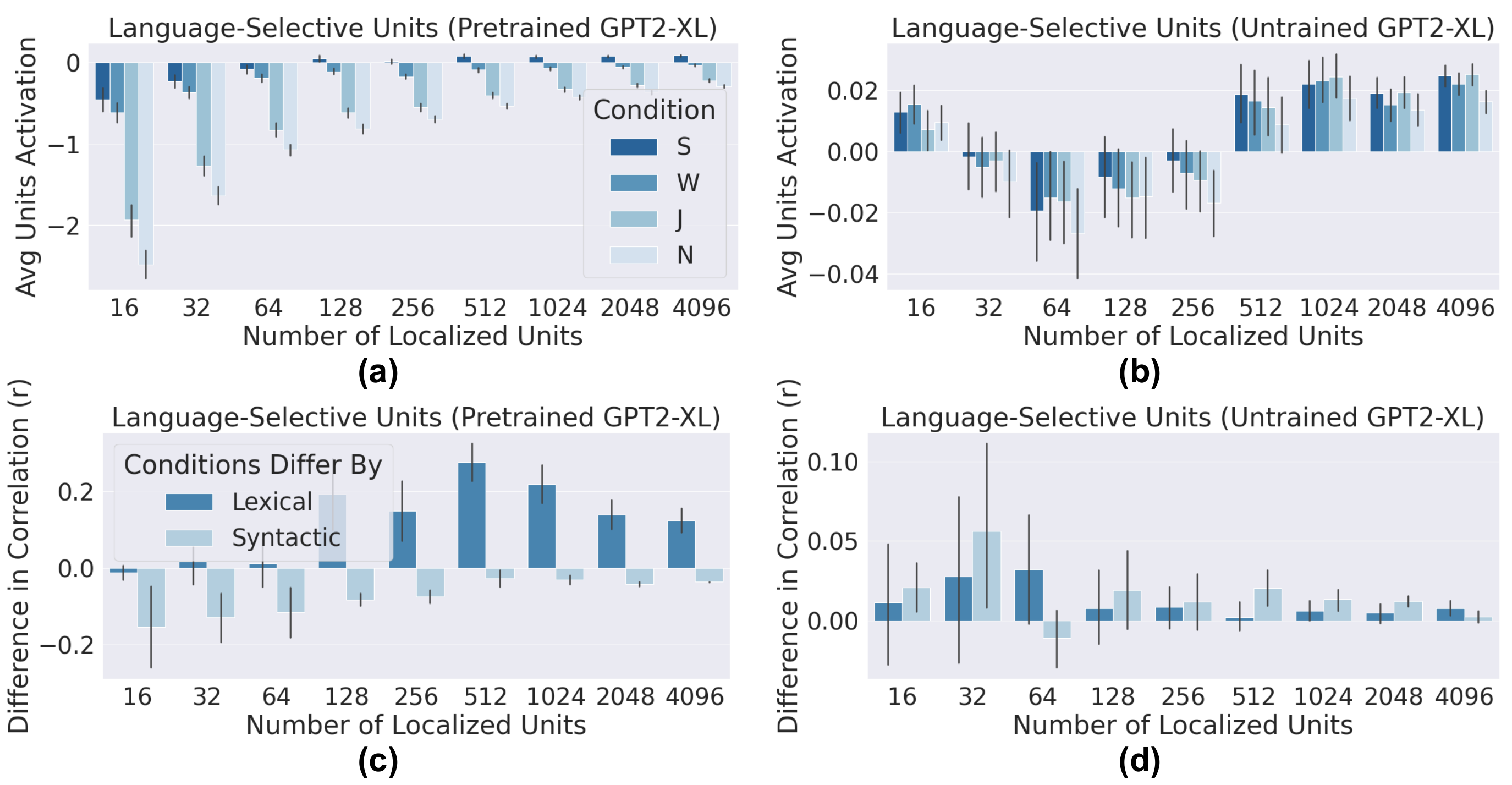}
    \caption{
        \textbf{Univariate Condition-Level Responses and Multivariate Representational Pattern Analysis using Pretrained and Untrained \modelname{GPT2-XL}}
        Here, we observe the effect of localizing different number of units on the univariate \textbf{(a,b)} and multivariate \textbf{(c,d)} representational analyses described in Section \ref{sec:hls-response-profiles}. We use 5 different initializations for the untrained models \textbf{(b,d)} for each value of $k$. Similar to Figure \ref{fig:neural-analyses} the error bars shows the variance across the stimuli and models evaluated.
    }
    \label{fig:neural-analyses-num-units}
\end{figure}

Figure \ref{fig:neural-analyses-num-units} repeats the same analyses conducted in Section \ref{sec:hls-response-profiles} but for different values of $k \in \{16, 32, 64, ..., 4096\}$ using pretrained and untrained \modelname{GPT2-XL}. For the pretrained model, we observe a response profile similar to what has been observed in the human language system across all values of $k$. Even when localizing only 16 units, the extracted representations exhibit similar selectivity for different conditions and can more reliably differentiate between lexical and non-lexical inputs compared to syntactic and non-syntactic inputs. However, for the untrained models, a large number of localized units is necessary to identify patterns akin to those found in the brain.

\section{Localized Units}
\label{app:localization-viz}

\begin{figure}
    \centering
    \includegraphics[width=1\linewidth]{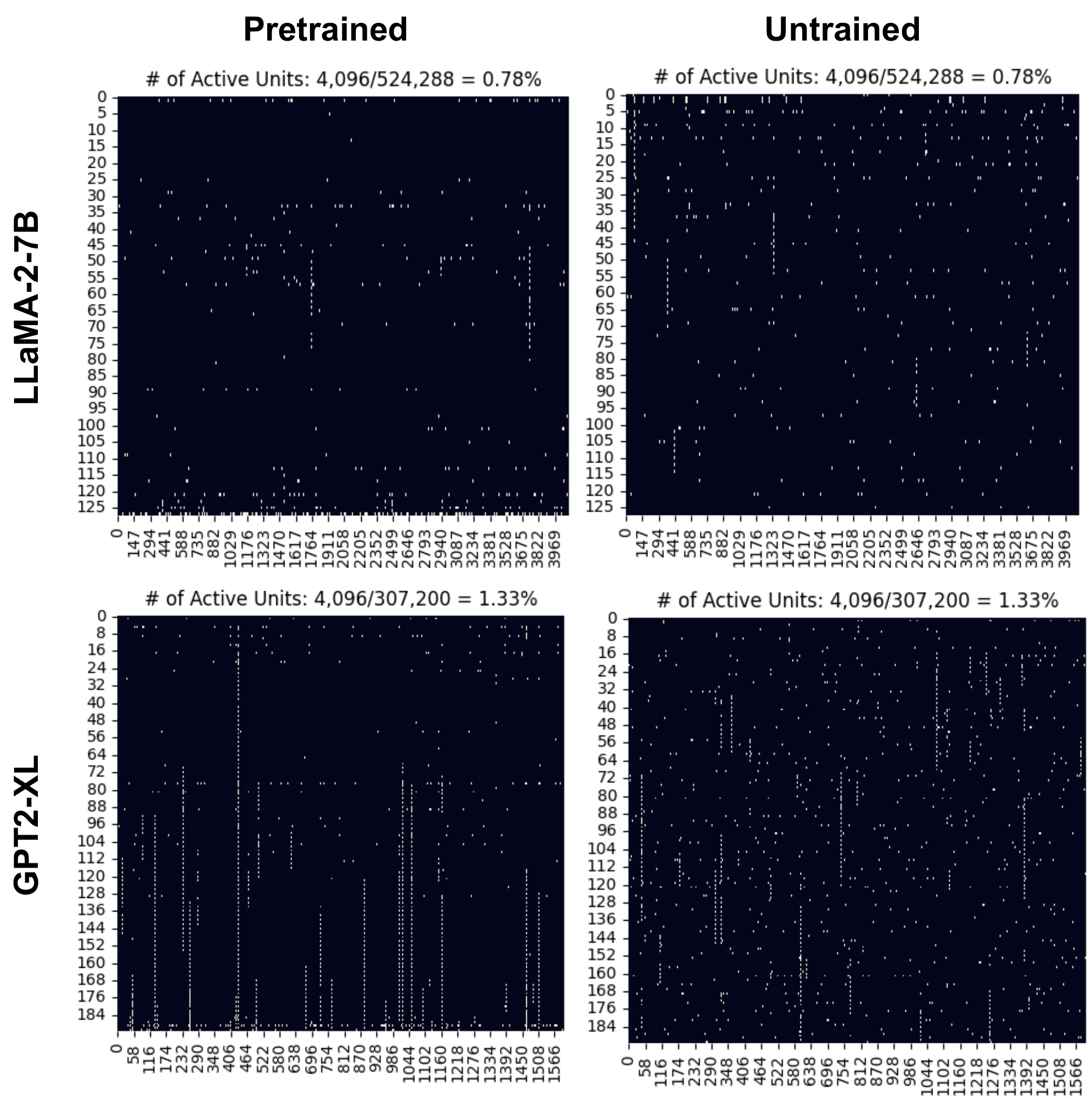}
    \caption{\textbf{Language Units Visualization}.
    Visualization of the top 4096 language selective units for the trained and an untrained versions of \modelname{LLaMA-2-7B} (top) and \modelname{GPT2-XL} (bottom). The depth is the vertical axis and goes from top to bottom, while the width is the horizontal axis.}
    \label{fig:language-masks}
\end{figure}

Figure \ref{fig:language-masks} shows a visualization of the top 4096 language selective units for \modelname{LLaMA-2-7B} and \modelname{GPT2-XL}. The vertical axes illustrates the depth, where every 4 ticks correspond to the components considered in a single Transformer block in the order in which they appear in the architecture. For instance, the first row in each matrix correspond to the output activations of the first layer normalization in the first Transformer block, this is then followed by the output of the multihead attention, then the second layer normalization then finally the output of the MLP. This sequence is repeated equal $L$ times where $L$ is the number of Transformer blocks in the corresponding architecture. \modelname{LLaMA-2-7B} has 32 blocks whereas \modelname{GPT2-XL} has 48 blocks, corresponding to 128 and 192 rows respectively. The horizontal axis is the hidden state dimension, which is 1600 for \modelname{GPT2-XL} and 4096 for \modelname{LLaMA-2-7B}. 

The language-selective units in the pretrained models appear to be more structured, where most of the units are at the last layers. The untrained ones on the other hand are more distributed around the network with little apparent structure.

\section{Training Setup}
\label{app:training-setup}

We train 6 different models on \code{wikiText-103-v1} available on HuggingFace datasets.\footnote{\url{https://huggingface.co/datasets/wikitext}} Each model was trained for 5 epochs with an effective batch-size of 128. We used a context-window size of 512 tokens. The learning rate started with a warm-up for 500 optimization steps to $5e^{-3}$ when training one Transformer block, and $5e^{-4}$ when training two blocks, then decayed linearly over the course of training. Model evaluation on the validation set was done every 1000 steps, and we took the checkpoint with the lowest validation loss.

\section{Experiments Compute Resources}

We used NVIDIA TITAN X GPUs to evaluate the brain alignment of the \modelname{GPT2} model family as well as for the shallow untrained architectures like \ourmodel. For the larger models such as those belonging to the \modelname{LLaMA-2} family we used NVIDIA A100 with 80 GBs of memory. The latter GPU was also used for training all the language modeling experiments.  

\end{document}